\documentclass[twoside]{article}
\usepackage[accepted]{aistats2025}
\usepackage[style=authoryear, natbib=true]{biblatex}
\usepackage{hyperref}
\usepackage{url}

\usepackage[utf8]{inputenc} 
\usepackage[T1]{fontenc}    
\usepackage{hyperref}       
\usepackage{url}            
\usepackage{booktabs}       
\usepackage{amsfonts}       
\usepackage{nicefrac}       
\usepackage{microtype}      
\usepackage{xcolor}         
\usepackage{pdfpages}
\usepackage{enumerate}
\usepackage{listings}
\definecolor{codegreen}{rgb}{0,0.6,0}
\definecolor{codegray}{rgb}{0.5,0.5,0.5}
\definecolor{codepurple}{rgb}{0.58,0,0.82}
\definecolor{backcolour}{rgb}{0.95,0.95,0.92}
\lstdefinestyle{mystyle}{
    backgroundcolor=\color{backcolour},   
    commentstyle=\color{codegreen},
    keywordstyle=\color{magenta},
    numberstyle=\tiny\color{codegray},
    stringstyle=\color{codepurple},
    basicstyle=\ttfamily\footnotesize,
    breakatwhitespace=false,         
    breaklines=true,                 
    captionpos=b,                    
    keepspaces=true,                 
    numbers=left,                    
    numbersep=5pt,                  
    showspaces=false,                
    showstringspaces=false,
    showtabs=false,                  
    tabsize=2
}
\usepackage{hyperref}
\usepackage{url}
\usepackage{booktabs}
\usepackage{multirow}
\usepackage{amsthm}
\newtheorem{definition}{Definition}

\usepackage{tabularx}
\usepackage{longtable}
\usepackage{array}
\usepackage{wrapfig}

\usepackage[T1]{fontenc}
\usepackage{lmodern}

\usepackage{amsmath,amsfonts}
\usepackage{booktabs}
\usepackage{rotating} 
\usepackage{lipsum} 
\usepackage{array}
\usepackage{multirow}
\usepackage[most]{tcolorbox}

\usepackage{enumitem} 
\hypersetup{
    colorlinks = true,
    linkcolor = blue,
    anchorcolor = blue,
    citecolor = black,
    filecolor = blue,
    urlcolor = blue
}

\usepackage{graphicx}
\usepackage[ruled,vlined]{algorithm2e}
\usepackage{caption}
\usepackage{subcaption}
\usepackage{pifont}
\newcommand{\cmark}{\ding{51}}%
\definecolor{lightgray}{gray}{0.75}
\newcommand{\xmark}{\textcolor{lightgray}{\ding{55}}}
\definecolor{economist}{RGB}{115,00,00} 
\definecolor{customgreen}{RGB}{116, 154, 114}
\definecolor{lightgreen}{RGB}{240, 246, 232}
\definecolor{greylight}{RGB}{242, 242, 242}
\definecolor{greydark}{RGB}{179, 179, 179}
\definecolor{ForestGreen}{RGB}{34, 139, 34}
\definecolor{pastelblue}{RGB}{46, 95, 127} 
\definecolor{pastelorange}{RGB}{201, 171, 102} 
\definecolor{pastelgreen}{RGB}{76, 124, 49} 
\newtcolorbox{greycustomblock}{
  colframe=greydark,        
  colback=greylight,        
  boxrule=1pt,              
  left=2.5pt,               
  right=3pt,                
  top=5pt,                  
  bottom=3pt,               
  arc=0pt,                  
  breakable,                
  before skip=0.2\baselineskip, 
  after skip=0.2\baselineskip,  
  left skip=0pt,            
  right skip=0pt,           
  enhanced jigsaw,          
  frame hidden,             
  overlay={                 
    \draw[greydark, line width=2pt]
      ([yshift=-1pt]frame.north west) -- ([yshift=1pt]frame.south west); 
  },
  fontupper=\fontfamily{lmr}\selectfont, 
}
\usepackage{fontawesome}
\usepackage{adjustbox} 
\usepackage{tikz}
\usepackage{float}

\usepackage[framemethod=TikZ]{mdframed}

\newcommand{\lightbulbicon}{%
  \begin{tikzpicture}[baseline=-0.5ex]
    \draw[fill=white, draw=customgreen, thick] (0,0) circle (1.5ex);
    \node[scale=0.8, color=customgreen] at (0,0) {\faLightbulbO~};
  \end{tikzpicture}%
}

\newtcolorbox{customblockquote}{
  colframe=customgreen,
  colback=lightgreen,
  boxrule=0pt,
  leftrule=2pt, 
  left=1pt,  
  right=3pt,
  top=5pt,
  bottom=3pt,
  arc=0pt,
  breakable,
  before skip=1.1\baselineskip,
  after skip=0.7\baselineskip,
  left skip=0pt,
  right skip=0pt,
  enhanced jigsaw,
  frame hidden,
   overlay={
    \draw[customgreen, line width=2pt] 
      (frame.north west) -- (frame.south west);
    \node[inner sep=0pt] at ([xshift=0pt, yshift=-1.3pt]frame.north west) {\lightbulbicon};
  },
  fontupper=\fontfamily{lmr}\selectfont,
  boxsep=3pt,
}

\usepackage[utf8]{inputenc}
\usepackage[english]{babel}
\usepackage{amssymb,amsmath,amsthm}
\usepackage{xcolor}
\usepackage{algorithmic}
\usepackage{tikz}
\usepackage{multirow}
\usepackage{multicol}

\newcommand{\mycomment}[1]{}
\DeclareMathOperator*{\argmin}{arg\,min}

\usepackage{mathtools} 

\begin{document}

\twocolumn[

\aistatstitle{Visualizing token importance for black-box language models}

\aistatsauthor{Paulius Rauba$^*$ \And Qiyao Wei$^*$ \And Mihaela van der Schaar}
\aistatsaddress{ University of Cambridge \And  University of Cambridge \And University of Cambridge }]

\begin{abstract}
We consider the problem of auditing \textit{black-box} large language models (LLMs) to ensure they behave reliably when deployed in production settings, particularly in high-stakes domains such as legal, medical, and regulatory compliance. Existing approaches for LLM auditing often focus on isolated aspects of model behavior, such as detecting specific biases or evaluating fairness. We are interested in a more general question---can we understand how the outputs of black-box LLMs depend on \textit{each input token}? There is a critical need to have such tools in real-world applications that rely on inaccessible API endpoints to language models. However, this is a highly non-trivial problem, as LLMs are stochastic functions (i.e. two outputs will be different by chance), while computing prompt-level gradients to approximate input sensitivity is infeasible. To address this, we propose Distribution-Based Sensitivity Analysis (DBSA), a lightweight model-agnostic procedure to evaluate the sensitivity of the output of a language model for each input token, without making any distributional assumptions about the LLM. DBSA is developed as a \textit{practical tool} for practitioners, enabling quick, plug-and-play visual exploration of LLMs reliance on specific input tokens. Through illustrative examples, we demonstrate how DBSA can enable users to inspect LLM inputs and find sensitivities that may be overlooked by existing LLM interpretability methods.
\end{abstract}

\vspace{-4mm}
\section{Introduction}

\vspace{-2mm}

Auditing black-box language models is a fundamental requirement to ensuring they behave reliably when deployed in the real-world. Society implicitly expects LLMs to operate ethically, comply with regulations, and deliver technically sound responses without causing harm. This has resulted in the development of broad audit frameworks focused on governance and regulatory oversight \citep{mokander2023auditing, mesko2023imperative, raji2022outsider}. Despite these high-level frameworks, \textit{practical algorithms} that practitioners could employ to audit LLMs are scarce.

\textbf{Why is there a need for practical LLM audit algorithms?} While high-level frameworks are necessary, there is an increasing demand for \textit{practical} algorithms that provide detailed, interpretable insights into LLM behavior. The urgency for such tools arises from two factors: \textit{mitigating real-world harm} and \textit{complying with regulatory demands}. Without precise auditing tools, LLMs may propagate biases or deliver inconsistent outputs, leading to societal harm \citep{gehman2020realtoxicityprompts, nozza2021honest} or diverge from their intended purpose over time, causing unforseen negative impacts \citep{lauer2021you, rahwan2018society, dafoe2018ai}. Regulatory frameworks like the EU AI Act are considering classifying LLMs as ``high-risk AI systems'' that would require mandatory third-party audits before deployment \citep{helberger2023chatgpt, mokander2023auditing}. Existing tools do not meet these needs because they either focus narrowly on specific biases or require complex, benchmark-oriented setups that are not adaptable to diverse real-world scenarios. 

\textbf{What is missing from current LLM auditing methods?} The auditing landscape for LLMs is dominated by methods that target isolated dimensions, such as bias detection \citep{borkan2019nuanced, dixon2018measuring, park2018reducing}, fairness evaluation \citep{garg2019counterfactual}, or adversarial robustness \citep{geisler2024attacking}. While these approaches are valuable for targeted analysis, they lack the flexibility and general applicability required by practitioners who need quick, interpretable feedback on LLM behavior across varied domains. What is missing is a practical, easy-to-deploy, statistically-grounded tool that provides actionable, visual insights into the behavior of any black-box LLM.

\textbf{How can sensitivity analysis serve as a practical and interpretable tool for auditing LLMs?} We diverge from existing auditing mechanisms and define an entirely new task---developing a model-agnostic framework for LLM sensitivity analysis. Sensitivity analysis offers a practical and interpretable way to audit LLMs by showing how model outputs change with variations in input. In a nutshell, the primary question that sensitivity analysis for language models answers is the following:

\vspace{-3mm}
\begin{quote}
\textit{``How do the answers produced by a language model depend on each input token?''}
\end{quote}
\vspace{-3mm}

Such an approach allows practitioners to visualize and interpret the influence of each token, providing a model-agnostic way of auditing LLM behavior. For example, Table \ref{tab:example} shows how sensitivity analysis can reveal unwanted model behavior when evaluating legal prompts, providing clear, actionable insights that standard benchmarks might overlook. \textit{The ability to identify such context-dependent sensitivities allows practitioners to explore if model behavior is appropriate for the given scenario}.

\begin{table}[]
    \centering
    \caption{\textbf{Why sensitivity analysis is important for model auditing}. Two examples show sensitivity scores for each word when an LLM is asked for legal advice. Darker shades indicate higher importance. In Example 1, the model focuses on the alleged crime, while in Example 2, its sensitivity to the person’s name indicates an undesired reliance when the name is irrelevant to the LLM response. Example 2 could prompt further investigation in practical scenarios. \\ \textbf{Legend}:  \raisebox{0.5ex}{\protect\colorbox{red!100}{\rule{0ex}{0.5ex}\hspace{0.5ex}}} Most important \hspace{2ex} \raisebox{0.5ex}{\protect\colorbox{red!7}{\rule{0ex}{0.5ex}\hspace{0.5ex}}} Least important}
    \vspace{-5mm}
    \label{tab:example}
    \vspace{2ex}
    \begin{tabular}{@{}p{0.49\textwidth}@{}}
        \toprule
        \textbf{Example 1 (Desired behavior):} \\
        \footnotesize
        \colorbox{red!10}{John} \ \colorbox{red!10}{Doe} \ is \ \colorbox{red!5}{accused} \ of \ \colorbox{red!70}{murder}.
        \\[2ex]
        \hline
        \vspace{-1ex}
        \textbf{Example 2 (Undesired behavior):} \\
        \footnotesize
        \colorbox{red!70}{John} \ \colorbox{red!70}{Doe} \ is \ \colorbox{red!5}{accused} \ of \ \colorbox{red!10}{murder}.\\
        \bottomrule
    \end{tabular}
    \vspace{-3mm}
\end{table}

However, implementing such a tool poses unique challenges. Repurposing existing gradient-based methods for black-box LLMs \citep{zeiler2014visualizing} is impossible because we assume no access to LLM internals. Another alternative could be to manually change the input to a desired perturbation and inspect the changes. However, this is problematic because LLMs are stochastic generators, i.e. two outputs will vary by chance. Therefore, it is \textit{insufficient} to look at two individual answers due to the stochastic nature of the LLMs; one must consider the \textit{entire distribution of possible answers} rather than individual outputs.

\textbf{Our solution}. To provide a practical solution for these challenges, we introduce \textit{Distribution-Based Sensitivity Analysis} (DBSA). DBSA is a lightweight, exploratory tool that finds the most important tokens in an input prompt. It is designed to be immediately usable by practitioners without requiring access to proprietary model internals or specialized knowledge. To achieve this, we approximate sensitivity analysis by considering each token's nearest neighbors in embedding space, performing Monte Carlo sampling for each neighbor and comparing Monte Carlo estimates on a reduced similarity space. With this formulation, we (i) capture the stochastic nature of LLM outputs more faithfully than point estimates; (ii) connect LLM outputs to distributional testing; (iii) quantify effect size; (iv) maintain model and input agnosticism, and (v) allow to directly visualize and highlight tokens based on their importance.

\textbf{What practical value does DBSA offer?} DBSA is designed for immediate application across various domains such as legal, medical, and customer service, where understanding token-level behavior is important. Existing methods, such as gradient-based sensitivity and bias evaluation techniques, are limited to specific use cases or require access to model internals. DBSA addresses \textit{the entirely new task of generalized sensitivity analysis for black-box LLMs}, applicable across diverse domains. Because this is a new interpretability setup, it does not lend itself naturally to benchmark comparisons---a reason why we focus on illustrative examples and ablation studies in Sec. \ref{sec:case_studies} - \ref{sec:insights}. We see DBSA as having direct applicability in many applications, such as visualizing all tokens in a prompt (Table \ref{tab1:highlights}), top \textit{n} tokens and computing effect sizes (Table \ref{tab2:highlights}), and others.


\begin{customblockquote}
    \textbf{Contributions}. \textbf{\textcolor{pastelgreen}{\textcircled{1}}} \textit{Conceptually}, we introduce an entirely new task---auditing how the output distribution of black-box LLMs depend on specific input tokens (Sec. \ref{sec:limitations}). \textbf{\textcolor{pastelgreen}{\textcircled{2}}} \textit{Technically}, we develop DBSA, a lightweight, easily deployable model-agnostic algorithm that uses finite-sample approximations and statistical testing to measure the impact of token-level changes (Sec. \ref{sec:finite_approx} - \ref{sec:DBSA}). \textbf{\textcolor{pastelgreen}{\textcircled{3}}} \textit{Empirically}, we show how DBSA offers black-box interpretability (Sec. \ref{sec:case_studies}) and build trust in the method by performing relevant quantitative ablations (Sec. \ref{sec:insights}).
\end{customblockquote}

\vspace{-4mm}
\section{Unique challenges of obtaining token importance scores in black-box language models}
\label{sec:limitations}

We start by defining our problem (Sec. \ref{subsec:problem_formulation}) and explaining why evaluating LLM responses requires looking at \textit{distributions} of answers (Sec. \ref{subsec:hypothesis_testing}). We then explore the unique challenges this poses (Sec. \ref{subsec:challenges}), proposing a procedure to address them (Sec. \ref{sec:finite_approx}).

\vspace{-2mm}
\subsection{Problem formulation}
\label{subsec:problem_formulation}

Let $\mathcal{X}$ denote the input space, where each input $x \in \mathcal{X}$ is a sequence of words or tokens $x = (t_1, t_2, \dots, t_n)$. Let $\mathcal{Y}$ denote the output space of the machine learning system. We define the system as a stochastic mapping $\mathcal{S}: \mathcal{X} \rightarrow \mathcal{P}(\mathcal{Y})$, where $\mathcal{P}(\mathcal{Y})$ denotes the space of probability distributions over $\mathcal{Y}$.

\textbf{Objective}. Our objective is to address the following research question: Given a black-box ML system $\mathcal{S}$, an input $x \in \mathcal{X}$, and an individual token(s) $t_i$ at position $i$ in $x$, how can we systematically measure and interpret the sensitivity of the output distribution $\mathcal{S}(x)$ with respect to small perturbations of $t_i$? We measure such \textit{minimal} changes in token-level sensitivity as follows. 

\begin{definition}[Token-Level Sensitivity]
\label{def:token_sensitivity}

The token-level sensitivity of the machine learning system $\mathcal{S}$ at position $i$ in input $x \in \mathcal{X}$ is defined as the expected change in the output distribution $\mathcal{S}(x)$ resulting from a small perturbation $\delta t_i$ to the word $t_i$, normalized by the magnitude of the perturbation:

\[
\text{Sensitivity}(t_i) = \lim_{\|\delta t_i\| \rightarrow 0} \frac{\omega\left( \mathcal{S}(x),\; \mathcal{S}\left(x_{[t_i \leftarrow t_i + \delta t_i]}\right) \right)}{\|\delta t_i\|}
\]

$x_{[t_i \leftarrow t_i + \delta t_i]}$ denotes the input sequence where $t_i$ has been perturbed by $\delta t_i$ and $\omega$ is a suitable distance measure.

\end{definition}

\subsection{Assessing sensitivity through comparison of output distributions}
\label{subsec:hypothesis_testing}

How can we evaluate token-level sensitivity? This requires (a) analyzing how the entire distribution of responses would change under the defined sensitivity metric; (b) understanding how to approximate such small perturbations in a discrete token space and a black-box setting; (c) evaluate how to take into account the inherent stochasticity of language model responses. We address these in turn.

First, we need a framework to evaluate how the \textit{entire distribution of responses} would change under a particular change in the input. To do this, we take the approach of using Monte Carlo estimates of perturbed inputs as a measure of constructing such a distribution \citep{rauba2024quantifying}. Suppose we swap an input token with its nearest neighbor in embedding space. A naive approach might involve comparing individual outputs from $\mathcal{S}$ for the original input $x$ and its perturbed version $x_{[t_i \leftarrow t_i + \delta t_i]}$:

\vspace{-3mm}
\[
y = \mathcal{S}(x), \quad y' = \mathcal{S}\left(x_{[t_i \leftarrow t_i + \delta t_i]}\right)
\]

However, since $\mathcal{S}$ produces random outputs even for the same input, such a comparison between single samples $y$ and $y'$ is insufficient. Any observed difference could merely be a consequence of the system's stochasticity rather than the effect of the perturbation $\delta t_i$.

To address this challenge, we employ a statistical approach that compares the entire output distributions of $\mathcal{S}$ for the original and perturbed inputs \citep{rauba2024quantifying}. Specifically, we consider the distributions $\mathcal{S}(x)$ and $\mathcal{S}\left(x_{[t_i \leftarrow t_i + \delta t_i]}\right)$ over the output space $\mathcal{Y}$. Therefore, our objective is to determine whether the swapped nearest neighbor token $\delta t_i$ leads to a significant change in the output distribution. This can be formalized as a hypothesis testing problem:

\vspace{-5mm}
\[\begin{aligned}    H_0 &: \mathcal{S}(x) = \mathcal{S}\left(x_{[t_i \leftarrow t_i + \delta t_i]}\right) \quad \text{(No effect)} \\
    H_1 &: \mathcal{S}(x) \neq \mathcal{S}\left(x_{[t_i \leftarrow t_i + \delta t_i]}\right) \quad \text{(Has an effect)}
\end{aligned}
\]
\vspace{-3mm}

The primary benefit of such a distributional formulation is that it captures the full stochastic behavior of $\mathcal{S}$ instead of just a single realization. This means we could perform statistical inference by directly comparing these distributions and understanding how much the outputs have shifted across the whole output space, even detecting subtle shifts that might not be apparent from individual samples. However, while theoretically grounded, the usage of such a hypothesis testing framework in the context of LLMs poses unique practical challenges that do not appear in regular settings. 

\subsection{Challenges with analyzing token-level output distributions}
\label{subsec:challenges}

Building on the foundational work of \citet{rauba2024quantifying}, we identify three key challenges in analyzing output distributions for token-level sensitivity, with particular emphasis on the discrete nature of nearest-neighbor perturbations.

$\blacktriangleright$ \textbf{Challenge 1: Computational intractability}. As established in prior work, the fundamental computational barrier lies in the exponential size of the output space $\mathcal{Y}$. For language models operating on vocabulary $V$ with sequence length $L$, we have $|\mathcal{Y}| = |V|^L$ possible outputs. The probability of any output sequence $y = (y_1, y_2, \dots, y_L) \in \mathcal{Y}$ under input $x$ is given by:

\[
\mathcal{S}(x)(y) = P(y \mid x) = \prod_{t=1}^{L} P(y_t \mid y_{<t}, x)
\]

where $y_{<t} = (y_1, \dots, y_{t-1})$ represents the token history. Comparing distributions $\mathcal{S}(x)$ and $\mathcal{S}(x_{[t_i \leftarrow t_i + \delta t_i]})$ requires evaluating these probabilities across the entire output space - a computation that grows exponentially with both vocabulary size and sequence length.

$\blacktriangleright$ \textbf{Challenge 2: Semantic interpretability}. Furthermore, statistical differences between output distributions may not reflect meaningful semantic changes. That is, while the outputs might be different token-wise, they might convey the same semantic meaning. Consider the outputs:

$y_1$: ``\textit{Targeted radiation therapy is suggested}'', $y_2$: ``\textit{We suggest targeted radiation therapy}''

While $\mathcal{S}(x)(y_1) \neq \mathcal{S}(x)(y_2)$ in general, both convey identical medical recommendations. Therefore, we require a way to evaluate outputs such that we take into account whether the \textit{semantic} meaning has changed.
 
$\blacktriangleright$ \textbf{Challenge 3: Discrete token space constraints}. Third, we require to perform swaps in discrete token space. The definition of token-level sensitivity in Definition~\ref{def:token_sensitivity} assumes the existence of infinitesimal perturbations $\delta t_i$. However, in a discrete token space without access to embeddings, such continuous perturbations are undefined. Instead, we must work with discrete jumps between tokens, where the smallest possible perturbation is a substitution with the nearest neighbor:

\[
\min_{\delta t_i} \|\delta t_i\| = \min_{t_j \in \mathcal{N}(t_i)} \text{dist}(t_i, t_j)
\]

where $\mathcal{N}(t_i)$ denotes the set of nearest neighbors of token $t_i$.

In order to develop a method to visualize tokens, we must make it computationally tractable, evaluate the effect of sensitivity on a semantic level and take into account the discrete nature of the token space. The following sections introduce an approach that addresses this.

\begin{customblockquote}
\textbf{Takeaway}. The discrete nature of nearest-neighbor token perturbations introduces fundamental constraints on sensitivity analysis that require extending existing distributional approaches to handle non-infinitesimal changes in the black-box LLM setting.
\end{customblockquote}

\section{Token-level sensitivity with finite sample approximations}
\label{sec:finite_approx}

How can we address the challenges outlined in Section~\ref{subsec:challenges}? We propose a light-weight finite sample approximation procedure that enables practical estimation of token-level sensitivity through hypothesis testing on output distributions.

\vspace{-2mm}
\subsection{Addressing outlined challenges}

$\blacktriangleright$ \textbf{Addressing challenge 1: Computational intractability}. We approximate the intractable distributions $\mathcal{S}(x)$ and $\mathcal{S}(x_{[t_i \leftarrow t_i']})$ using finite samples obtained via Monte Carlo sampling of the LLM. We generate $n$ independent samples from each distribution, which is used to calculate the token-level sensitivities.

\begin{align*}
\hat{\mathcal{S}}(x) &= \{ y^{(1)}, y^{(2)}, \dotsc, y^{(n)} \}, \quad y^{(j)} \sim \mathcal{S}(x), \\
\hat{\mathcal{S}}(x_{[t_i \leftarrow t_i']}) &= \{ y'^{(1)}, y'^{(2)}, \dotsc, y'^{(n)} \}, \quad y'^{(j)} \sim \mathcal{S}(x_{[t_i \leftarrow t_i']}).
\end{align*}

$\blacktriangleright$ \textbf{Addressing challenge 2: Semantic interpretability}. Recall that statistical differences between output distributions may not reflect meaningful semantic changes. To ensure that we are indeed measuring semantic distances, we define a semantic similarity function $s: \mathcal{Y} \times \mathcal{Y} \rightarrow [0,1]$ that quantifies the semantic similarity between outputs. We use a similarity metric in Sec. \ref{sec:DBSA} to evaluate distributional differences.

$\blacktriangleright$. \textbf{Addressing challenge 3: Discrete token space constraints)}. We approximate the infinitesimal perturbation $\delta t_i$ by substituting $t_i$ with one of its $k$ nearest neighbors in the embedding space. Let $\phi: \mathcal{W} \rightarrow \mathbb{R}^d$ be a token embedding function, and let $\mathcal{N}_k(t_i)$ denote the set of $k$ nearest neighbors of $t_i$ based on the similarity metric $s$. We select a small perturbation $t_i' \in \mathcal{N}_k(t_i)$ and define the perturbation magnitude as $\|\delta t_i\| = \left\| \phi(t_i') - \phi(t_i) \right\|$.

\begin{figure}[]
    \centering
    \includegraphics[width=\linewidth]{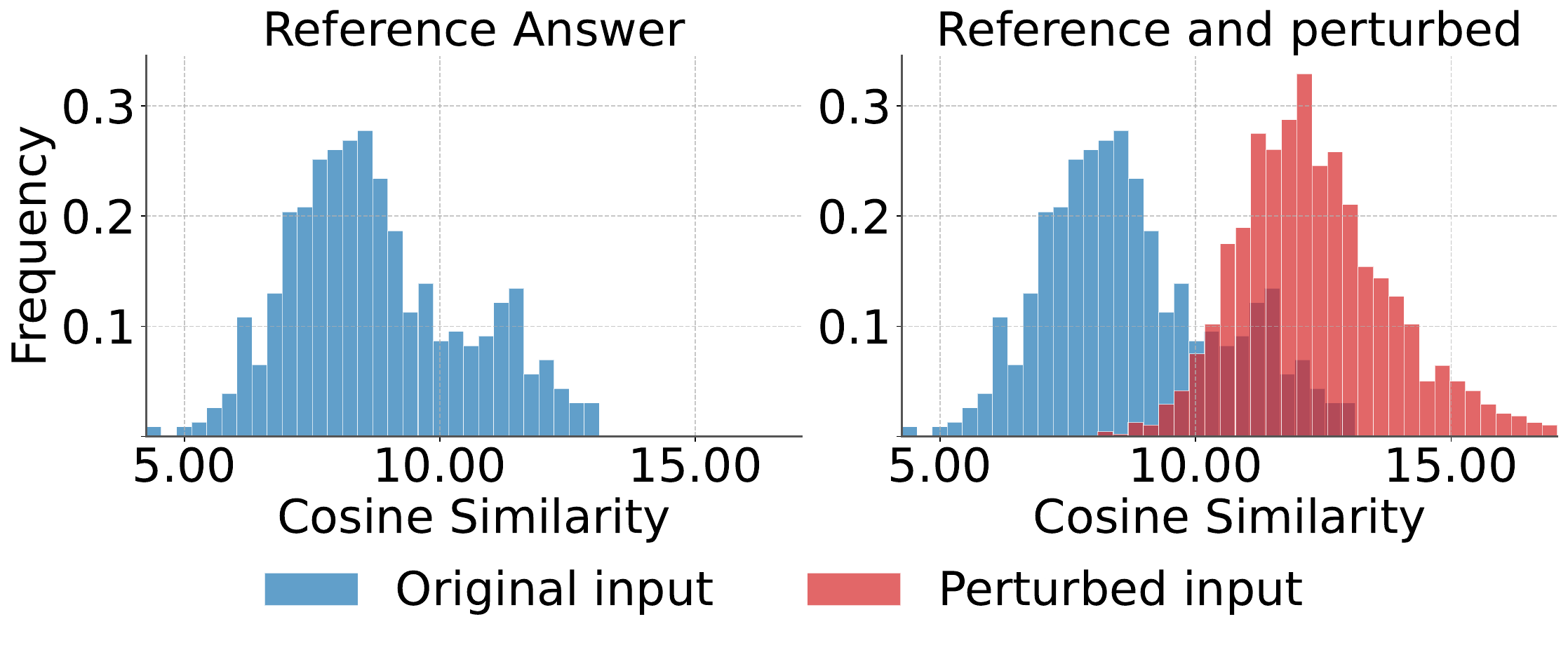}
    \caption{\textbf{Example of null and alternative distributions.} The null distribution $P_0$ (left, blue) is constructed based on the intrinsic variability of responses. The alternative distribution with a perturbed input of a \textit{single} nearest neighbor $P_1$ (right, red) is quantified with respect to the original distributions. This figure shows the output distribution change in the cosine similarity space.}
    \label{fig:null_distribution}
    \vspace{-2mm}
    \rule{\linewidth}{.5pt}
    \vspace{-7mm}

\end{figure}

\subsection{Quantifying token-level sensitivity}

Let $P_0$ be the distribution of similarities between pairs within $\hat{\mathcal{S}}(x)$, and $P_1$ be the distribution of similarities between pairs from $\hat{\mathcal{S}}(x)$ and $\hat{\mathcal{S}}(x_{[t_i \leftarrow t_i']})$:

\begin{align*}
P_0 &= \left\{ s\left(y^{(i)}, y^{(j)}\right) \mid 1 \leq i, j \leq n \right\}, \\
P_1 &= \left\{ s\left(y^{(i)}, y'^{(j)}\right) \mid 1 \leq i, j \leq n \right\},
\end{align*}

where \( y^{(i)}, y^{(j)} \overset{\text{i.i.d.}}{\sim} \mathcal{S}(x) \) and \( y'^{(j)} \overset{\text{i.i.d.}}{\sim} \mathcal{S}(x_{[t_i \leftarrow t_i']} ) \).Here, $P_0$ captures the intrinsic variability within the original output (equivalent to the \textit{null distribution}), whereas $P_1$ captures the cross-distribution similarities between the original and swapped token outputs (equivalent to the \textit{alternative distribution}), c.f. Figure \ref{fig:null_distribution}, where each swapped token is a nearest neighbor of the original token. We have therefore constructed two distributions which represent the variability in answer similarities as a proxy for sensitivity (Def. \ref{def:token_sensitivity}). In Sec. \ref{sec:DBSA}, we show how such distributions can be used to obtain token-level sensitivities and associated p-values.

Our hypothesis test is then formulated as:

\vspace{-4mm}
\begin{align*}
H_0 &: \mathbb{E}[P_0] = \mathbb{E}[P_1], \quad \text{(No significant effect)}, \\
H_1 &: \mathbb{E}[P_0] \neq \mathbb{E}[P_1], \quad \text{(Perturbation shifts similarity)}.
\end{align*}

We estimate the token-level sensitivity $\mathcal{S}$ at position $i$ by computing the average difference between the null distribution and alternative distribution.

\textbf{Advantages}. With this formulation, we (i) capture the stochastic nature of LLM outputs more faithfully than point estimates; (ii) connect LLM outputs to frequentist hypothesis testing; (iii) quantify effect size; and (iv) maintain model and input agnosticism.

\vspace{-2mm}
\begin{customblockquote}
\textbf{Takeaway}. Using finite-sample approximations can address the three primary challenges associated with evaluating token-level sensitivity.
\end{customblockquote}

\section{Distribution-based sensitivity analysis}
\label{sec:DBSA}
\vspace{-2mm}
Now, we present a light model-agnostic methodology for assessing the token-level sensitivity of LLMs. Our approach avoids restrictive distributional assumptions about LLM outputs. We enable frequentist statistical hypothesis testing using effect size and p-values through the construction of null and alternative distributions. Importantly, our framework is applicable to \textit{any perturbation} and \textit{any language model}, with the minimal requirement of being able to sample from the language model's output distribution and construct embeddings. We outline the general procedure below and provide other implementation details in Appendix \ref{sec:implementation_details}.

\textbf{DBSA in a nutshell}. DBSA computes token-level sensitivity by measuring the effect of token perturbations on LLM output distributions. Given an input sequence \( x \) and its unique tokens \( T \), indices \( \mathcal{I}_w \) for each token \( w \) are identified, and for each \( i \in \mathcal{I}_w \), we find the \( k \) nearest neighbors \( \mathcal{N}_k(t_i) \) of \( t_i \) in the embedding space using a distance function \( D \). For each neighbor \( t_i' \in \mathcal{N}_k(t_i) \), a perturbed sequence \( x^{(t_i')} \) is generated by replacing \( t_i \) with \( t_i' \). The LLM produces samples \( Y = \{y^{(j)}\}_{j=1}^n \sim \mathcal{S}(x) \) and \( Y' = \{y'^{(j)}\}_{j=1}^m \sim \mathcal{S}(x^{(t_i')}) \), and their embeddings \( \Phi \) and \( \Phi' \) are used to compute pairwise similarities with a function \( s \).

\textbf{Distance measure}. There are many possible distances metrics which might offer different properties. We seek for a distance which would (i) directly compute pairwise similarities; (ii) avoid intermediate density estimation; (iii) capture different moments of the distribution; (iv) would directly compute p-values, and (v) scale well with the number of samples. A natural choice for this is the \textit{energy distance}. The energy distance between \( \hat{\mathcal{S}}(x) \) and \( \hat{\mathcal{S}}(x^{(t_i')}) \) is calculated as \( E(\hat{\mathcal{S}}(x), \hat{\mathcal{S}}(x^{(t_i')})) = 2A - B - C \), where \( A = \frac{1}{nm} \sum_{a=1}^n \sum_{b=1}^m s(\phi(y^{(a)}), \phi(y'^{(b)})) \), \( B = \frac{1}{n^2} \sum_{a=1}^n \sum_{b=1}^n s(\phi(y^{(a)}), \phi(y^{(b)})) \), and \( C = \frac{1}{m^2} \sum_{a=1}^m \sum_{b=1}^m s(\phi(y'^{(a)}), \phi(y'^{(b)})) \). With this distance, sensitivity scores \( \omega_{wi}^{(w'_i)} \) and p-values \( p_{wi}^{(w'_i)} \) are computed for each token position and neighbor, then averaged across neighbors and instances to yield final scores \( \omega_w \) and \( p_w \) for each token \( w \). This model-agnostic approach only requires the ability to sample outputs and compute embeddings, ensuring broad applicability.

The chosen distance measure is a design choice and not a necessary component of the sensitivity analysis more broadly. One can choose a different $\omega$, a different number of nearest neighbors or embedding components to suit each use case separately.

We provide more implementation details in Appendix \ref{sec:implementation_details}, including an algorithm for DBSA and a discussion on different design choices. We perform some major ablations of such choices in Sec. \ref{sec:insights} and Appendix \ref{sec:extended_experiments}.

\vspace{-2mm}
\begin{customblockquote}
\textbf{Takeaway}. A distribution-based approach provides a lightweight solution that captures the full variability of LLM outputs, enabling practitioners to audit model behavior without requiring access to internals.
\end{customblockquote}

\vspace{-2mm}
\section{Illustrative examples}
\label{sec:case_studies}
\vspace{-2mm}

\textbf{Purpose and scope}. In this section, we present two illustrative examples that demonstrate how DBSA can be used as a decision-support tool in real-world domains. Importantly, DBSA can be used \textit{in addition to} other audit mechanisms to aid decision support. Our examples are based on text-based prompts because, unlike many common LLM interpretability methods, we do not assume any ground-truth labels. All experimental details can be found in Appendix \ref{sec:experimental_setting}\footnote{Associated code for DBSA can be found here: https://github.com/vanderschaarlab/visualizing-token-importance}.

\textbf{Comparison to other auditing algorithms}. Existing sensitivity methods cannot be directly repurposed for these use cases due to the unique problem formulation (Sec. \ref{sec:limitations}) that requires understanding \textit{token-level} impact on LLMs.

\vspace{-2mm}
\subsection*{Illustrative Example I: Legal Systems}

\begin{greycustomblock}
\textbf{Example I}. Using DBSA in legal audit

\vspace{-2.5mm}
\rule{\linewidth}{.5pt}
\vspace{-3mm}

\textbf{The Problem}. A tech company uses an AI-powered system to analyze legal contracts for potential risks by querying a language model for advice. Recently, the system missed a critical clause that resulted in massive legal damage. 

\textbf{The Need}. The company cannot rely on black-box LLMs for contract evaluation. It needs more transparency and currently has no way of auditing models. They want to implement an additional monitoring system which can help understand \textit{how} the answers might change if the information were slightly different.
\end{greycustomblock}

\textbf{Context}. The consequences of misinterpreting key contractual elements can be severe. In this example, we illustrate how DBSA can be integrated into the auditing process to provide \textit{token-level insights} that directly inform legal teams. Table \ref{tab1:highlights} provides an example to visualize DBSA.

\begin{table}[h]
    \centering
    \caption{\textbf{Example of using DBSA to audit legal documents for which words impact answers the most}. Each word is highlighted based on its normalized impact on the output distribution $\omega$. Darker color highlights greater impact on answers. \\ \textbf{Legend}:  \raisebox{0.5ex}{\protect\colorbox{red!100}{\rule{0ex}{0.5ex}\hspace{0.5ex}}} Most important \hspace{2ex} \raisebox{0.5ex}{\protect\colorbox{red!7}{\rule{0ex}{0.5ex}\hspace{0.5ex}}} Least important}
    \vspace{-2.5mm}
    \label{tab1:highlights}
    \begin{tabular}{p{0.47\textwidth}}
    \hline
    \rule{0pt}{2.5ex} 
    \tiny
\colorbox{red!18.75}{Company} \colorbox{red!0.0}{A} \colorbox{red!42.5}{agrees} \colorbox{red!0.0}{to} \colorbox{red!14.38}{pay} \colorbox{red!18.75}{Company} \colorbox{red!13.75}{B} \colorbox{red!0.0}{\$10} \colorbox{red!35.0}{million} \colorbox{red!14.69}{for} \colorbox{red!13.44}{developing} \colorbox{red!0.0}{a} \colorbox{red!86.67}{revolutionary} \colorbox{red!100.0}{AI} \colorbox{red!57.5}{software} \colorbox{red!18.44}{within} \colorbox{red!0.0}{12} \colorbox{red!14.06}{months} \colorbox{red!0.0}{.} \colorbox{red!6.25}{If} \colorbox{red!18.75}{Company} \colorbox{red!13.75}{B} \colorbox{red!5.0}{fails} \colorbox{red!0.0}{to} \colorbox{red!13.12}{deliver} \colorbox{red!0.0}{a} \colorbox{red!16.56}{fully} \colorbox{red!16.88}{functional} \colorbox{red!9.06}{product} \colorbox{red!19.38}{by} \colorbox{red!12.5}{the} \colorbox{red!80.0}{deadline} \colorbox{red!0.0}{,} \colorbox{red!15.94}{they} \colorbox{red!7.19}{must} \colorbox{red!93.33}{refund} \colorbox{red!0.0}{50} \colorbox{red!12.81}{\%} \colorbox{red!0.0}{of} \colorbox{red!12.5}{the} \colorbox{red!9.69}{payment} \colorbox{red!0.0}{and} \colorbox{red!8.12}{provide} \colorbox{red!17.81}{an} \colorbox{red!12.19}{additional} \colorbox{red!11.25}{3} \colorbox{red!14.06}{months} \colorbox{red!0.0}{of} \colorbox{red!15.62}{development} \colorbox{red!20.0}{at} \colorbox{red!10.31}{no} \colorbox{red!65.0}{extra} \colorbox{red!10.0}{cost} \colorbox{red!0.0}{.} \colorbox{red!10.94}{However} \colorbox{red!0.0}{,} \colorbox{red!50.0}{if} \colorbox{red!12.5}{the} \colorbox{red!18.12}{delay} \colorbox{red!17.19}{is} \colorbox{red!5.94}{due} \colorbox{red!0.0}{to} \colorbox{red!15.31}{circumstances} \colorbox{red!7.5}{beyond} \colorbox{red!18.75}{Company} \colorbox{red!13.75}{B} \colorbox{red!0.0}{'} \colorbox{red!15.0}{s} \colorbox{red!27.5}{reasonable} \colorbox{red!20.0}{control} \colorbox{red!0.0}{,} \colorbox{red!10.62}{these} \colorbox{red!19.69}{penalties} \colorbox{red!19.06}{shall} \colorbox{red!11.88}{not} \colorbox{red!8.75}{apply} \colorbox{red!0.0}{.} \colorbox{red!5.62}{This} \colorbox{red!72.5}{agreement} \colorbox{red!17.19}{is} \colorbox{red!8.44}{governed} \colorbox{red!19.38}{by} \colorbox{red!80.0}{California} \colorbox{red!6.88}{law} \colorbox{red!0.0}{and} \colorbox{red!16.25}{any} \colorbox{red!6.56}{disputes} \colorbox{red!19.06}{shall} \colorbox{red!5.31}{be} \colorbox{red!7.81}{resolved} \colorbox{red!9.38}{through} \colorbox{red!11.56}{binding} \colorbox{red!17.5}{arbitration} \colorbox{red!0.0}{.}

    \\[2ex] 

    \end{tabular}
\end{table}

\textbf{Discussion I.} DBSA ensures the LLM focuses on critical contractual terms like ``agreement'', ``California'', and ``AI software'' It also identifies irrelevant terms, such as ``product'' or ``governed'', confirming their minimal impact. This validates the model’s behavior, giving legal teams confidence in its outputs. Therefore, DBSA can be used to highlight how LLM answers depend on each word to corroborate expert knowledge and evaluate LLM safety. 

\vspace{-2mm}
\subsection*{Illustrative Example II: Clinical Support}

\begin{greycustomblock}
\textbf{Example II}. Using DBSA in clinical support

\vspace{-2.5mm}
\rule{\linewidth}{.5pt}
\vspace{-3mm}

\textbf{The Problem}. A hospital has implemented an LLM-powered clinical decision support system to assist doctors. The system analyzes patient data (e.g. medical history) to provide treatment recommendations. Recently, there was a near-miss incident where the system suggested an incorrect diagnosis that could have led to potentially harmful treatment. Upon review, it was discovered that \textit{a single word} in the patient's symptom description significantly altered the system output.

\textbf{The Need}. The hospital cannot use an LLM without having an additional mechanism to understand the impact of each word on the output of the model.
\end{greycustomblock}

\textbf{Context}. Decision-support systems must be interpretable and reliable, as their outputs directly impact patient safety. We can employ DBSA to identify the top $k$ words which change the output the most with minor variability and compute their effect sizes ($\omega$) as well as p-values. We show such an example in  Table \ref{tab2:highlights}.

\newcommand{\tightcolorbox}[2]{\colorbox{#1}{\kern-0.1em\relax#2\kern-0.25em\relax}}

\begin{table}[h]
    \centering
    \caption{\textbf{Example of using DBSA to identify top words affecting the output and identify effects of minor variations}. The top $k=5$ words are highlighted, and their effect size, p-value, and statistical significance (p < $\alpha$, where $\alpha = 0.05$) are computed. We see that none of the words affect the output significantly due to the high p-value. \\ \textbf{Legend}:  \raisebox{0.5ex}{\protect\colorbox{red!100}{\rule{0ex}{0.5ex}\hspace{0.5ex}}} Most important \hspace{2ex} \raisebox{0.5ex}{\protect\colorbox{red!7}{\rule{0ex}{0.5ex}\hspace{0.5ex}}} Least important}
    \vspace{-2.5mm}
    \label{tab2:highlights}
    \begin{tabular}{p{0.45\textwidth}}
    \hline
    \rule{0pt}{2.5ex} 
    \tiny
Patient is a 45-year-old male presenting with progressive dyspnea on exertion over the past two weeks. On \colorbox{red!80.0}{examination}, patient appears mildly distressed. \colorbox{red!90.0}{Lower} extremities show 2+ pitting edema to \colorbox{red!70.0}{mid} -shin bilaterally. Chest X-ray shows pulmonary vascular congestion. Clinical presentation is consistent with new-onset \colorbox{red!100.0}{congestive} heart failure, likely due to \colorbox{red!60.0}{hypertensive} heart disease.

\end{tabular}

\begin{tabular}{cccc}
    \toprule
    \textbf{Word} & \textbf{Effect size} & \textbf{p-value} & \textbf{p < $\alpha$} \\
    \midrule
    congestive & 0.08 & 0.11 & \xmark \\
    examination & 0.07 & 0.16 & \xmark \\
    Lower & 0.07 & 0.28 & \xmark \\
    mid & 0.07 & 0.39 & \xmark \\
    hypertensive & 0.06 & 0.31 & \xmark \\
    \bottomrule
\end{tabular}
\end{table}

\textbf{Discussion II}. DBSA confirms the LLM’s stability, as the highlighted words are appropriate and expected, demonstrating consistency in recommendations. In this way, DBSA can be used to highlight the top $k$ words and evaluate their effect size.

\begin{customblockquote}
\textbf{Takeaway}. Existing sensitivity methods are inadequate for highlighting the importance of each token. DBSA offers a model-agnostic solution to evaluate model sensitivity to each token.
\end{customblockquote}

\section{Insights Experiments}
\label{sec:insights}

\textbf{Purpose and scope}. To supplement the illustrative examples above, we aim to provide a preliminary analysis into the effectiveness and sensitivity of DBSA across various parameters and conditions. We do this by performing comprehensive empirical testing, ablating most of our design components and evaluating how they affect the results of DBSA. All experimental details can be found in Appendix \label{sec:experimental_setting}.

\vspace{-1mm}
\subsection{Experiment I: Effects across models}
\label{subsec:models}

\begin{figure}[h!]
    \centering
    \includegraphics[width=\linewidth]{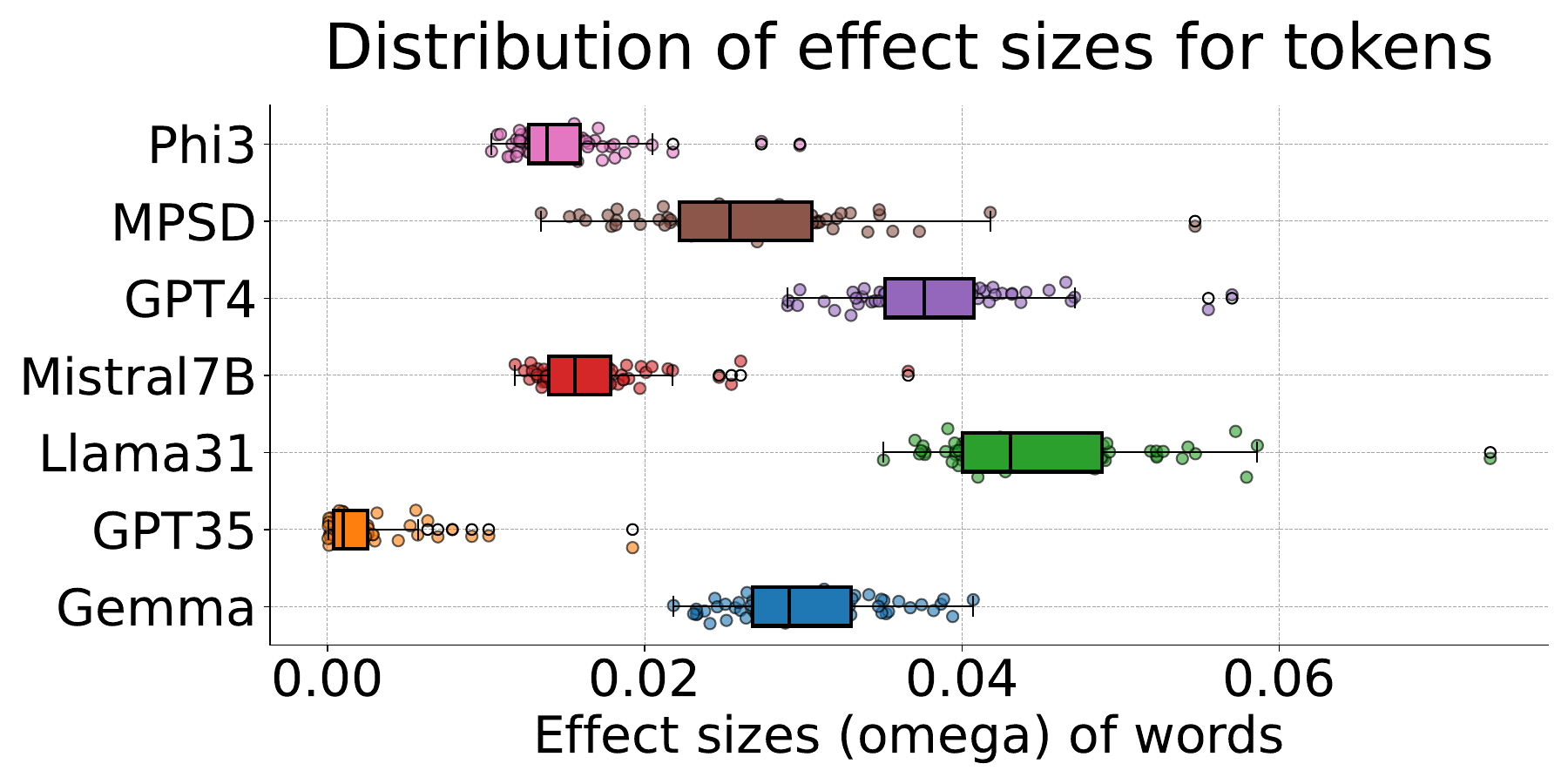}
    \caption{\textbf{Comparison of effect sizes across different language models}. The chart describes $\omega$ values for each word for seven large language models. The y-axis lists the models analyzed (full names in the Appendix, abbreviated for clarity). This comparison illustrates how the models vary in their sensitivity to input perturbations.}
    \label{fig:effect}
    \vspace{-2.4mm}
    \rule{\linewidth}{.5pt}
    \vspace{-2mm}
\end{figure}

\textcolor{pastelblue}{\textbf{Setup}}. We compare DBSA sensitivity analysis results across language models to assess consistency in identified sensitivities. Inputs are perturbed using nearest neighbors, and effect size distributions are evaluated across LLMs (see Figure \ref{fig:effect}) using the prompt from our clinical support example.

\textcolor{pastelgreen}{\textbf{Takeaway I}}. DBSA shows variation in sensitivity patterns across models, indicating that model architecture influences token importance. This validates DBSA’s capability to identify these variations.

\vspace{-2mm}
\subsection{Experiment II: Sensitivity across language models}

\textcolor{pastelblue}{\textbf{Setup}}. We compute Spearman rank correlation coefficients between models to quantify similarity in sensitivity rankings, assessing if models exhibit consistent behavior in token sensitivity. Results are in Table \ref{tab:spearman-rank}.

\begin{table}[h]
\centering
\caption{\textbf{Spearman Rank Coefficients between LLMs}. Higher values indicate stronger correlation in word rankings between models.}
\label{tab:spearman-rank}
\footnotesize
\resizebox{0.98\linewidth}{!}{\begin{tabular}{lccccccc}
\toprule\bf Model & \bf GPT-4 & \bf GPT-3.5 & \bf SmolLM & \bf \textbf{MagicPrompt} & \bf Mistral-7B & \bf Phi-3-mini & \bf Llama-3-8B \\
\midrule
GPT-4 & 1.00 & 0.14 & 0.03 & -0.07 & -0.02 & 0.04 & -0.07 \\
GPT-3.5 & 0.14 & 1.00 & 0.03 & 0.09 & 0.38 & -0.07 & 0.17 \\
SmolLM & 0.03 & 0.03 & 1.00 & 0.16 & 0.25 & 0.05 & 0.23 \\
MagicPrompt & -0.07 & 0.09 & 0.16 & 1.00 & 0.44 & 0.43 & 0.38 \\
Mistral-7B & -0.02 & 0.38 & 0.25 & 0.44 & 1.00 & 0.16 & 0.25 \\
Phi-3-mini & 0.04 & -0.07 & 0.05 & 0.43 & 0.16 & 1.00 & 0.11 \\
Llama-3-8B & -0.07 & 0.17 & 0.23 & 0.38 & 0.25 & 0.11 & 1.00 \\
\bottomrule
\end{tabular}}
\end{table}

\textcolor{pastelgreen}{\textbf{Takeaway II}}. The analysis reveals substantial variability in sensitivity patterns across models, with some showing higher alignment (e.g., models from similar architectures like GPT-4 and GPT-3.5) while others demonstrate divergent sensitivity behaviors. DBSA effectively identifies these discrepancies, offering a reliable tool for practitioners to evaluate model robustness and make informed decisions when considering model replacements or optimizations.

\vspace{-2mm}
\subsection{Experiment III: Evaluating similarity functions}

\begin{figure}[h!]
    \centering
    \includegraphics[width=\linewidth]{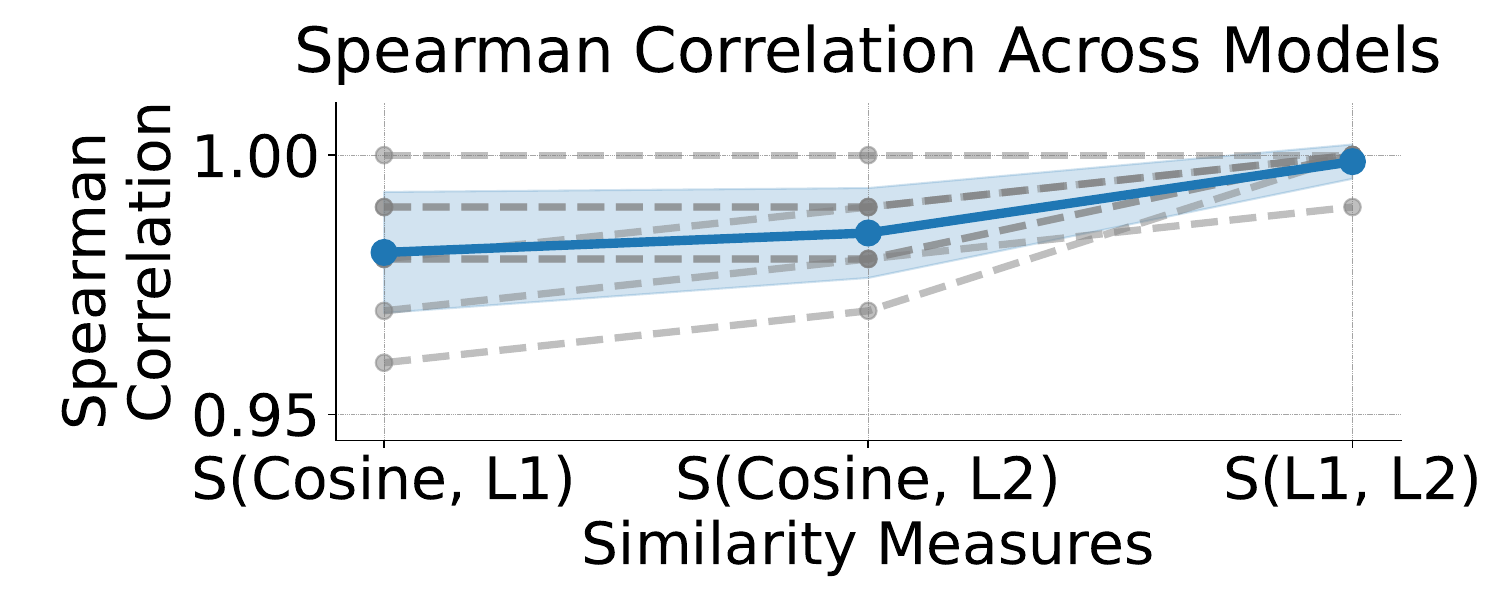}
    \caption{\textbf{Comparison of similarity functions (Cosine, L1, L2) in DBSA}. We show the average Spearman correlation across models (blue) and individual models (grey). Results indicate all three measures yield consistent outcomes, demonstrating their interchangeability within the DBSA framework.}
    \label{fig:cosine}
    \vspace{-2mm}
    \rule{\linewidth}{.5pt}
    \vspace{-2.5mm}
\end{figure}

\textcolor{pastelblue}{\textbf{Setup}}. We test three similarity metrics (Cosine, L1, L2) to verify if DBSA's sensitivity analysis is consistent across metrics. We measure the Spearman correlation of token $\omega$ scores across LLMs to ensure comparable results (see Fig, \ref{fig:cosine}).

\textcolor{pastelgreen}{\textbf{Takeaway III}}. All three similarity metrics produce similar results, demonstrating DBSA’s robustness. This confirms that practitioners can choose any standard metric without compromising the quality of analysis, offering flexibility in deploying DBSA.

\vspace{-2mm}
\subsection{Experiment IV: Evaluating the effect of Monte Carlo samples}

\begin{figure}
    \centering
    \includegraphics[width=\linewidth]{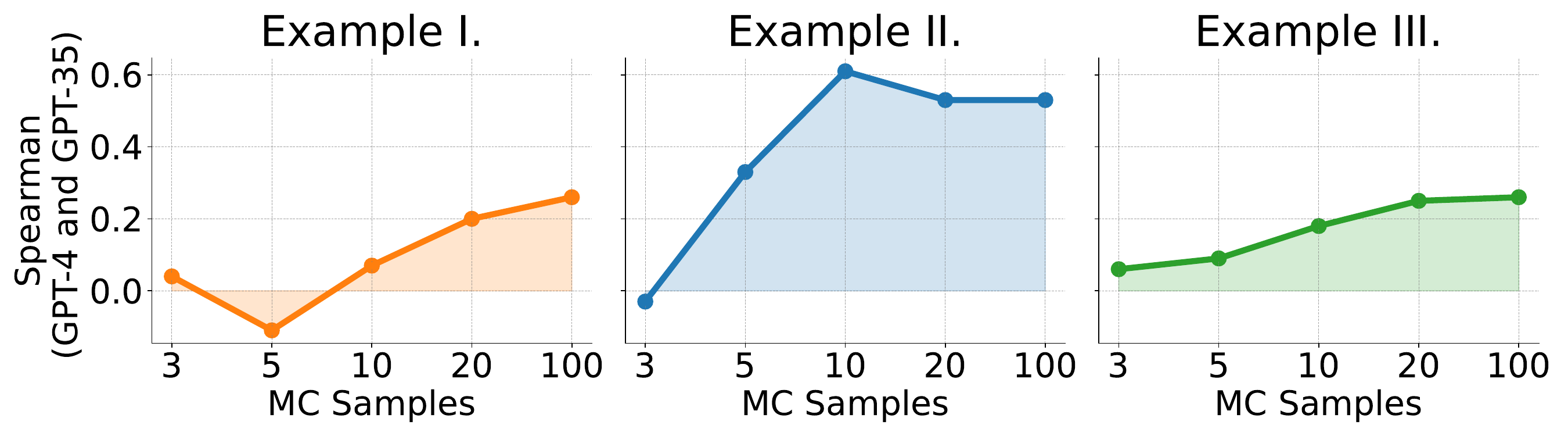}
    \caption{\textbf{Stabilization of Spearman correlation between GPT-3.5 and GPT-4 with increasing Monte Carlo (MC) samples for three prompts.} The figure shows that with fewer MC samples, the Spearman correlation fluctuates, indicating variability. As MC samples increase, the correlation stabilizes, showing greater reliability.}
    \label{fig:mc_sample}
    \vspace{-2mm}
    \rule{\linewidth}{.5pt}
    \vspace{-6mm}
\end{figure}

\textcolor{pastelblue}{\textbf{Setup}}. We assess how varying Monte Carlo (MC) sample sizes affects DBSA’s stability and reliability. We measure Spearman correlation between model outputs (e.g., GPT-4, GPT-3.5) to determine if larger sample sizes stabilize results across three prompts (Fig. \ref{fig:mc_sample}).

\begin{table*}[h!]
\centering
\caption{\textbf{Comparison to related work}. DBSA addresses the entirely new task of generalized sensitivity analysis for black-box LLMs for which there are no ground-truth labels. This differs from other sensitivity- or interpretability-based methods. \textbf{Abbreviations:} \textbf{(I)}: Usable on any black-box model; \textbf{(II)}: Enables statistical inference; \textbf{(III)}: Computes the effect size; \textbf{(IV)}: Assumption-free}
\tiny
\label{tab:llm_evaluation_methods}
\resizebox{\textwidth}{!}{%
\begin{tabular}{@{}p{0.13\textwidth}p{0.25\textwidth}ccccp{0.25\textwidth}@{}}
\toprule
\textbf{Method} & \textbf{Example Works} & \textbf{(I)} & \textbf{(II)} & \textbf{(III)} & \textbf{(IV)} & \textbf{Representative Question} \\
\midrule
Gradient Methods & \citep{geisler2024attacking} & \xmark & \cmark & \xmark & \cmark & How does model output change under infinitesimal perturbation? \\
\addlinespace
Measuring Unintended Bias & \citep{borkan2019nuanced, dixon2018measuring, park2018reducing} & \cmark & \cmark & \cmark & \xmark & Does this model have unintended biases in certain subgroups? \\
\addlinespace
Counterfactual Fairness & \citep{garg2019counterfactual} & \xmark & \xmark & \cmark & \xmark & How would the prediction change if the sensitive attribute were different? \\
\addlinespace
Text Summarization & \citep{bhandari2020re, zhang2019bertscore, zhao2019moverscore, lin2004rouge} & \xmark & \xmark & \cmark & \cmark & How well is this text summarized? \\
\hline
Sensitivity analysis for LLMs & Our work (DBSA) & \cmark & \cmark & \cmark & \cmark & Do the responses change if we change any input in the prompt? If so, how? \\
\bottomrule
\end{tabular}%
}
\vspace{-2mm}
\end{table*}

\textcolor{pastelgreen}{\textbf{Takeaway IV}}. As MC samples increase, DBSA’s outputs stabilize across larger models, confirming the importance of sample size for accuracy. This shows that DBSA’s reliability can be tuned based on sample quantity, offering guidelines for practitioners. This experiment establishes that DBSA provides consistent

\vspace{-2mm}
\subsection{Other empirical insights}
We perform evaluation on other components, such as performing ablations by evaluating different distance functions (Energy, Earth Mover's Distance, or mean distance of cosine similarities) and the effect of choosing a different number of nearest neighbors (Appendix \ref{sec:extended_experiments}).

\vspace{-2mm}
\section{Related work}
\label{sec:related_work}

Existing methods for evaluating the sensitivity and behavior of large language models (LLMs) primarily include gradient-based approaches, bias measurement, counterfactual fairness, and text summarization metrics. \textbf{Gradient-based methods} \citep{geisler2024attacking} require access to model internals, making them impractical for black-box models, and focus on performance rather than interpretability. \textbf{Bias easurement techniques }\citep{borkan2019nuanced, dixon2018measuring, park2018reducing} quantify fairness but are limited to specific subgroups and require human annotation, which restricts their flexibility and applicability to arbitrary perturbations. \textbf{Counterfactual fairness} approaches \citep{garg2019counterfactual} commonly rely on specific assumptions, labeled data, and fundamentally answer a different question, i..e. that of fairaness. \textbf{Text summarization metrics} such as ROUGE and BERTScore \citep{bhandari2020re, zhang2019bertscore, zhao2019moverscore, lin2004rouge} assess text generation quality but are not designed for inspecting output variability for input perturbations.

DBSA is a tangential auditing framework to the above-mentioned frameworks which answers a \textit{fundamentally different question}. Because this paper defines an entirely new task---that of developing a model-agnostic framework for token-level sensitivity --- we encourage practitioners to use DBSA \textit{in addition to} any existing techniques that are used to audit LLM models. This is useful, as DBSA supports statistical inference, computes effect sizes, and allows exploration of any input perturbation, offering a human-interpretable approach that prioritizes qualitative understanding over benchmark performance. This is also important, as there have been direct calls for practical mechanisms to audit language models in addition to existing methods \cite{mokander2023auditing, mesko2023imperative}. Table \ref{tab:llm_evaluation_methods} summarizes these methods across five criteria. Extended related work can be found in Appendix \ref{sec:extended_related_work}.

\vspace{-2mm}
\section{Discussion}
\vspace{-1mm}

\textbf{The need for practical algorithms to audit LLMs}. There is a growing need for \textit{practical} algorithms that can be used to audit LLMs. DBSA is a versatile, model-agnostic framework designed for practitioners who wish to understand how sensitive the model is to information in the prompt. DBSA addresses a new task within the LLM auditing literature that other methods have so far overlooked. We see DBSA functioning as a diagnostic instrument that \textit{complements} existing frameworks, especially in high-stakes settings. DBSA has immediate practical relevance to many practitioners who lack practical tools to audit LLMs. With this, we hope our work lays the foundation for a new class of statistics-based auditing tools.

\textbf{Limitations}. DBSA requires access to an embedding function and the ability to sample from the LLM multiple times which could limit its accessibility to some practitioners. We outline token-level sensitivity challenges in Sec. \ref{sec:limitations}. While our solution in Sec. \ref{sec:finite_approx} addresses these, it has limitations, such as finding reliable nearest neighbors (which may vary in magnitude in embedding space) or selecting appropriate distance metrics/embedding functions. Finally, we wish to acknowledge the importance of ensuring that DBSA is deployed responsibly and does not inadvertently enable malicious actors to exploit vulnerabilities in LLMs.

\textbf{Directions for Future Work} (1) DBSA could be extended to have a multi-level approach to interpretability. We think this is a non-trivial (but extremely fruitful) extension of our current work. Such an extension would have to address at least five different non-trivial challenges: (i) which levels are worth testing (multiple tokens? semantic meaning? sentence-level?); (ii) the mutual information between levels and their interactions (we cannot assume independence; otherwise, this extension loses its meaning); (iii) dealing with higher variance estimators (due to the non-zero covariances between levels); (iv) dealing with semantic drift (i.e. change in the meaning of the input across multiple places); and (v) repurposing the highlighting methodology (e.g. our current work offering highlights of tokens is extremely intuitive --- how can this be achieved otherwise?). (2) Variance decomposition. Token probabilities could facilitate a formal decomposition of variability in output distributions, akin to an ANOVA-style partitioning of variance. Specifically, this could disentangle within-prompt variability (changes due to perturbations within a given input) from between-prompt variability (differences arising from distinct inputs). Such an approach could quantify the relative contributions of perturbations at various levels—be it tokens, phrases, or broader input features—to the overall stochastic behavior of the model. (3) Bayesian sensitivity. Token probabilities could be used within a Bayesian framework, where they serve as priors to model the effects of input perturbations. By updating these priors with observed outputs, DBSA could estimate posterior distributions, providing a probabilistic characterization of how input modifications alter the likelihood of specific outputs.

\newpage
\printbibliography

@inproceedings{sikdar2021integrated,
  title={Integrated directional gradients: Feature interaction attribution for neural NLP models},
  author={Sikdar, Sandipan and Bhattacharya, Parantapa and Heese, Kieran},
  booktitle={Proceedings of the 59th Annual Meeting of the Association for Computational Linguistics and the 11th International Joint Conference on Natural Language Processing (Volume 1: Long Papers)},
  pages={865--878},
  year={2021}
}

@article{enguehard2023sequential,
  title={Sequential Integrated Gradients: a simple but effective method for explaining language models},
  author={Enguehard, Joseph},
  journal={arXiv preprint arXiv:2305.15853},
  year={2023}
}

@article{yuksekgonul2024textgrad,
  title={TextGrad: Automatic" Differentiation" via Text},
  author={Yuksekgonul, Mert and Bianchi, Federico and Boen, Joseph and Liu, Sheng and Huang, Zhi and Guestrin, Carlos and Zou, James},
  journal={arXiv preprint arXiv:2406.07496},
  year={2024}
}

@inproceedings{borkan2019nuanced,
  title={Nuanced metrics for measuring unintended bias with real data for text classification},
  author={Borkan, Daniel and Dixon, Lucas and Sorensen, Jeffrey and Thain, Nithum and Vasserman, Lucy},
  booktitle={Companion proceedings of the 2019 world wide web conference},
  pages={491--500},
  year={2019}
}

@inproceedings{dixon2018measuring,
  title={Measuring and mitigating unintended bias in text classification},
  author={Dixon, Lucas and Li, John and Sorensen, Jeffrey and Thain, Nithum and Vasserman, Lucy},
  booktitle={Proceedings of the 2018 AAAI/ACM Conference on AI, Ethics, and Society},
  pages={67--73},
  year={2018}
}

@article{park2018reducing,
  title={Reducing gender bias in abusive language detection},
  author={Park, Ji Ho and Shin, Jamin and Fung, Pascale},
  journal={arXiv preprint arXiv:1808.07231},
  year={2018}
}

@inproceedings{garg2019counterfactual,
  title={Counterfactual fairness in text classification through robustness},
  author={Garg, Sahaj and Perot, Vincent and Limtiaco, Nicole and Taly, Ankur and Chi, Ed H and Beutel, Alex},
  booktitle={Proceedings of the 2019 AAAI/ACM Conference on AI, Ethics, and Society},
  pages={219--226},
  year={2019}
}

@article{mohri2024language,
  title={Language models with conformal factuality guarantees},
  author={Mohri, Christopher and Hashimoto, Tatsunori},
  journal={arXiv preprint arXiv:2402.10978},
  year={2024}
}

@misc{lundbergunified,
  title={A unified approach to interpreting model predictions. NIPS’17: Proceedings of the 31st International Conference on Neural Information Processing Systems. December 2017 [Cited 2021 Jul 20]},
  author={Lundberg, SM and Lee, SI}
}

@inproceedings{sundararajan2017axiomatic,
  title={Axiomatic attribution for deep networks},
  author={Sundararajan, Mukund and Taly, Ankur and Yan, Qiqi},
  booktitle={International conference on machine learning},
  pages={3319--3328},
  year={2017},
  organization={PMLR}
}

@article{montavon2017explaining,
  title={Explaining nonlinear classification decisions with deep taylor decomposition},
  author={Montavon, Gr{\'e}goire and Lapuschkin, Sebastian and Binder, Alexander and Samek, Wojciech and M{\"u}ller, Klaus-Robert},
  journal={Pattern recognition},
  volume={65},
  pages={211--222},
  year={2017},
  publisher={Elsevier}
}

@inproceedings{ribeiro2016should,
  title={" Why should i trust you?" Explaining the predictions of any classifier},
  author={Ribeiro, Marco Tulio and Singh, Sameer and Guestrin, Carlos},
  booktitle={Proceedings of the 22nd ACM SIGKDD international conference on knowledge discovery and data mining},
  pages={1135--1144},
  year={2016}
}

@article{singh2024rethinking,
  title={Rethinking interpretability in the era of large language models},
  author={Singh, Chandan and Inala, Jeevana Priya and Galley, Michel and Caruana, Rich and Gao, Jianfeng},
  journal={arXiv preprint arXiv:2402.01761},
  year={2024}
}

@article{clark2019does,
  title={What Does Bert Look At? An Analysis of Bert’s Attention},
  author={Clark, Kevin},
  journal={arXiv preprint arXiv:1906.04341},
  year={2019}
}

@article{morris2023text,
  title={Text embeddings reveal (almost) as much as text},
  author={Morris, John X and Kuleshov, Volodymyr and Shmatikov, Vitaly and Rush, Alexander M},
  journal={arXiv preprint arXiv:2310.06816},
  year={2023}
}

@article{zou2023representation,
  title={Representation engineering: A top-down approach to ai transparency},
  author={Zou, Andy and Phan, Long and Chen, Sarah and Campbell, James and Guo, Phillip and Ren, Richard and Pan, Alexander and Yin, Xuwang and Mazeika, Mantas and Dombrowski, Ann-Kathrin and others},
  journal={arXiv preprint arXiv:2310.01405},
  year={2023}
}

@article{belrose2023eliciting,
  title={Eliciting latent predictions from transformers with the tuned lens},
  author={Belrose, Nora and Furman, Zach and Smith, Logan and Halawi, Danny and Ostrovsky, Igor and McKinney, Lev and Biderman, Stella and Steinhardt, Jacob},
  journal={arXiv preprint arXiv:2303.08112},
  year={2023}
}

@article{ghandeharioun2024patchscope,
  title={Patchscope: A unifying framework for inspecting hidden representations of language models},
  author={Ghandeharioun, Asma and Caciularu, Avi and Pearce, Adam and Dixon, Lucas and Geva, Mor},
  journal={arXiv preprint arXiv:2401.06102},
  year={2024}
}

@article{bhandari2020re,
  title={Re-evaluating evaluation in text summarization},
  author={Bhandari, Manik and Gour, Pranav and Ashfaq, Atabak and Liu, Pengfei and Neubig, Graham},
  journal={arXiv preprint arXiv:2010.07100},
  year={2020}
}

@article{zhang2019bertscore,
  title={Bertscore: Evaluating text generation with bert},
  author={Zhang, Tianyi and Kishore, Varsha and Wu, Felix and Weinberger, Kilian Q and Artzi, Yoav},
  journal={arXiv preprint arXiv:1904.09675},
  year={2019}
}

@article{zhao2019moverscore,
  title={MoverScore: Text generation evaluating with contextualized embeddings and earth mover distance},
  author={Zhao, Wei and Peyrard, Maxime and Liu, Fei and Gao, Yang and Meyer, Christian M and Eger, Steffen},
  journal={arXiv preprint arXiv:1909.02622},
  year={2019}
}

@inproceedings{lin2004rouge,
  title={Rouge: A package for automatic evaluation of summaries},
  author={Lin, Chin-Yew},
  booktitle={Text summarization branches out},
  pages={74--81},
  year={2004}
}

@article{mokander2023auditing,
  title={Auditing large language models: a three-layered approach},
  author={M{\"o}kander, Jakob and Schuett, Jonas and Kirk, Hannah Rose and Floridi, Luciano},
  journal={AI and Ethics},
  pages={1--31},
  year={2023},
  publisher={Springer}
}

@article{mesko2023imperative,
  title={The imperative for regulatory oversight of large language models (or generative AI) in healthcare},
  author={Mesk{\'o}, Bertalan and Topol, Eric J},
  journal={NPJ digital medicine},
  volume={6},
  number={1},
  pages={120},
  year={2023},
  publisher={Nature Publishing Group UK London}
}

@inproceedings{raji2022outsider,
  title={Outsider oversight: Designing a third party audit ecosystem for ai governance},
  author={Raji, Inioluwa Deborah and Xu, Peggy and Honigsberg, Colleen and Ho, Daniel},
  booktitle={Proceedings of the 2022 AAAI/ACM Conference on AI, Ethics, and Society},
  pages={557--571},
  year={2022}
}

@article{helberger2023chatgpt,
  title={ChatGPT and the AI Act},
  author={Helberger, Natali and Diakopoulos, Nicholas and others},
  journal={Internet Policy Review},
  volume={12},
  number={1},
  pages={1--6},
  year={2023},
  publisher={Alexander von Humboldt Institute for Internet and Society}
}

@article{gehman2020realtoxicityprompts,
  title={Realtoxicityprompts: Evaluating neural toxic degeneration in language models},
  author={Gehman, Samuel and Gururangan, Suchin and Sap, Maarten and Choi, Yejin and Smith, Noah A},
  journal={arXiv preprint arXiv:2009.11462},
  year={2020}
}

@inproceedings{nozza2021honest,
  title={HONEST: Measuring hurtful sentence completion in language models},
  author={Nozza, Debora and Bianchi, Federico and Hovy, Dirk and others},
  booktitle={Proceedings of the 2021 Conference of the North American Chapter of the Association for Computational Linguistics: Human Language Technologies},
  year={2021},
  organization={Association for Computational Linguistics}
}

@article{lauer2021you,
  title={You cannot have AI ethics without ethics},
  author={Lauer, Dave},
  journal={AI and Ethics},
  volume={1},
  number={1},
  pages={21--25},
  year={2021},
  publisher={Springer}
}

@article{rahwan2018society,
  title={Society-in-the-loop: programming the algorithmic social contract},
  author={Rahwan, Iyad},
  journal={Ethics and information technology},
  volume={20},
  number={1},
  pages={5--14},
  year={2018},
  publisher={Springer}
}

@article{dafoe2018ai,
  title={AI governance: a research agenda},
  author={Dafoe, Allan},
  journal={Governance of AI Program, Future of Humanity Institute, University of Oxford: Oxford, UK},
  volume={1442},
  pages={1443},
  year={2018}
}

@article{geisler2024attacking,
  title={Attacking large language models with projected gradient descent},
  author={Geisler, Simon and Wollschl{\"a}ger, Tom and Abdalla, MHI and Gasteiger, Johannes and G{\"u}nnemann, Stephan},
  journal={arXiv preprint arXiv:2402.09154},
  year={2024}
}

@inproceedings{zeiler2014visualizing,
  title={Visualizing and understanding convolutional networks},
  author={Zeiler, Matthew D and Fergus, Rob},
  booktitle={Computer Vision--ECCV 2014: 13th European Conference, Zurich, Switzerland, September 6-12, 2014, Proceedings, Part I 13},
  pages={818--833},
  year={2014},
  organization={Springer}
}

@article{webster2020measuring,
  title={Measuring and reducing gendered correlations in pre-trained models},
  author={Webster, Kellie and Wang, Xuezhi and Tenney, Ian and Beutel, Alex and Pitler, Emily and Pavlick, Ellie and Chen, Jilin and Chi, Ed and Petrov, Slav},
  journal={arXiv preprint arXiv:2010.06032},
  year={2020}
}

@article{gallegos2024bias,
  title={Bias and fairness in large language models: A survey},
  author={Gallegos, Isabel O and Rossi, Ryan A and Barrow, Joe and Tanjim, Md Mehrab and Kim, Sungchul and Dernoncourt, Franck and Yu, Tong and Zhang, Ruiyi and Ahmed, Nesreen K},
  journal={Computational Linguistics},
  pages={1--79},
  year={2024},
  publisher={MIT Press 255 Main Street, 9th Floor, Cambridge, Massachusetts 02142, USA~…}
}

@article{cowgill2017algorithmic,
  title={Algorithmic bias: A counterfactual perspective},
  author={Cowgill, Bo and Tucker, Catherine},
  journal={NSF Trustworthy Algorithms},
  volume={3},
  year={2017}
}

@article{gao2015exploratory,
  title={Exploratory visualization design towards online social network privacy and data literacy},
  author={Gao, Bo},
  year={2015}
}

@article{jung2018algorithmic,
  title={Algorithmic decision making in the presence of unmeasured confounding},
  author={Jung, Jongbin and Shroff, Ravi and Feller, Avi and Goel, Sharad},
  journal={arXiv preprint arXiv:1805.01868},
  year={2018}
}

@article{wen2024hard,
  title={Hard prompts made easy: Gradient-based discrete optimization for prompt tuning and discovery},
  author={Wen, Yuxin and Jain, Neel and Kirchenbauer, John and Goldblum, Micah and Geiping, Jonas and Goldstein, Tom},
  journal={Advances in Neural Information Processing Systems},
  volume={36},
  year={2024}
}

@article{wang2023robust,
  title={How robust is your fairness? evaluating and sustaining fairness under unseen distribution shifts},
  author={Wang, Haotao and Hong, Junyuan and Zhou, Jiayu and Wang, Zhangyang},
  journal={Transactions on machine learning research},
  volume={2023},
  year={2023},
  publisher={NIH Public Access}
}

@article{rauba2024quantifying,
  title={Quantifying perturbation impacts for large language models},
  author={Rauba, Paulius and Wei, Qiyao and van der Schaar, Mihaela},
  journal={arXiv preprint arXiv:2412.00868},
  year={2024}
}

@article{rauba2024context,
  title={Context-aware testing: A new paradigm for model testing with large language models},
  author={Rauba, Paulius and Seedat, Nabeel and Ruiz Luyten, Max and van der Schaar, Mihaela},
  journal={Advances in Neural Information Processing Systems},
  volume={37},
  pages={112505--112553},
  year={2024}
}

@article{rauba2024self,
  title={Self-healing machine learning: A framework for autonomous adaptation in real-world environments},
  author={Rauba, Paulius and Seedat, Nabeel and Kacprzyk, Krzysztof and van der Schaar, Mihaela},
  journal={Advances in Neural Information Processing Systems},
  volume={37},
  pages={42225--42267},
  year={2024}
}

@article{astorga2024active,
  title={Active learning with llms for partially observed and cost-aware scenarios},
  author={Astorga, Nicol{\'a}s and Liu, Tennison and Seedat, Nabeel and van der Schaar, Mihaela},
  journal={Advances in Neural Information Processing Systems},
  volume={37},
  pages={20819--20857},
  year={2024}
}

@article{hadi2023survey,
  title={A survey on large language models: Applications, challenges, limitations, and practical usage},
  author={Hadi, Muhammad Usman and Qureshi, Rizwan and Shah, Abbas and Irfan, Muhammad and Zafar, Anas and Shaikh, Muhammad Bilal and Akhtar, Naveed and Wu, Jia and Mirjalili, Seyedali and others},
  journal={Authorea Preprints},
  volume={3},
  year={2023}
}

@article{liu2024large,
  title={Large language models to enhance bayesian optimization},
  author={Liu, Tennison and Astorga, Nicol{\'a}s and Seedat, Nabeel and van der Schaar, Mihaela},
  journal={arXiv preprint arXiv:2402.03921},
  year={2024}
}

\section*{Checklist}
 \begin{enumerate}

 \item For all models and algorithms presented, check if you include:
 \begin{enumerate}
   \item A clear description of the mathematical setting, assumptions, algorithm, and/or model. [Yes] We include a clear description of the mathematical setting in Section 2, and appropriate algorithmic details.
   \item An analysis of the properties and complexity (time, space, sample size) of any algorithm. [Yes] We analyze challenges including computation complexity in Section 2.
   \item (Optional) Anonymized source code, with specification of all dependencies, including external libraries. [No] All source code will be made available upon publication.
 \end{enumerate}

 \item For any theoretical claim, check if you include:
 \begin{enumerate}
   \item Statements of the full set of assumptions of all theoretical results. [Yes] We include a clear statement of assumptions in Section 2.
   \item Complete proofs of all theoretical results. [Not Applicable]
   \item Clear explanations of any assumptions. [Yes] We include a clear explanation in Section 2
 \end{enumerate}

 \item For all figures and tables that present empirical results, check if you include:
 \begin{enumerate}
   \item The code, data, and instructions needed to reproduce the main experimental results (either in the supplemental material or as a URL). [Yes] We include the data used in this experiment, and we discuss the model design choices that are made. All source code and reproduction instructions will be made available upon publication.
   \item All the training details (e.g., data splits, hyperparameters, how they were chosen). [Yes] All experiment details are described in Section 5.
    \item A clear definition of the specific measure or statistics and error bars (e.g., with respect to the random seed after running experiments multiple times). [Yes] We clearly describe the functions and metrics we are using.
    \item A description of the computing infrastructure used. (e.g., type of GPUs, internal cluster, or cloud provider). [Yes] We describe further experiment settings, including the compute used, in the appendix.
 \end{enumerate}

 \item If you are using existing assets (e.g., code, data, models) or curating/releasing new assets, check if you include:
 \begin{enumerate}
   \item Citations of the creator If your work uses existing assets. [Yes] We appropriately document the models that we are using.
   \item The license information of the assets, if applicable. [Yes] We appropriately document the license information.
   \item New assets either in the supplemental material or as a URL, if applicable. [Yes] We have further specification in the supplementary materials.
   \item Information about consent from data providers/curators. [Not Applicable] We do not collect data from external sources.
   \item Discussion of sensible content if applicable, e.g., personally identifiable information or offensive content. [Not Applicable] There are no sensible contents in this paper.
 \end{enumerate}

 \item If you used crowdsourcing or conducted research with human subjects, check if you include:
 \begin{enumerate}
   \item The full text of instructions given to participants and screenshots. [Not Applicable] We do not use crowdsourcing or conducted research with human subjects.
   \item Descriptions of potential participant risks, with links to Institutional Review Board (IRB) approvals if applicable. [Not Applicable] We do not use crowdsourcing or conducted research with human subjects.
   \item The estimated hourly wage paid to participants and the total amount spent on participant compensation. [Not Applicable] We do not use crowdsourcing or conducted research with human subjects.
 \end{enumerate}

 \end{enumerate}

\newpage

\appendix

\onecolumn
\aistatstitle{Supplementary Materials}

\section*{Appendix Contents}
\begin{itemize}[left=0pt, label={}, itemsep=1.2em] 
    \item \textbf{\hyperref[sec:extended_related_work]{1. Extended Related Work}}
    \item \textbf{\hyperref[sec:implementation_details]{2. Implementation Details}}
    \item \textbf{\hyperref[sec:extended_experiments]{3. Extended Experiments}}
    \item \textbf{\hyperref[sec:experimental_setting]{4. Experimental Details}}
    \item \textbf{\hyperref[sec:prompt_examples]{5. Prompt Examples}}
\end{itemize}
\newpage

\section{Extended related work}
\label{sec:extended_related_work}

\subsection{Feature Attribution}
\label{sec:appendix-feature-attribution}

The most relevant statistical methods for interpretability fall under "Feature attribution", such as SHAPLEY values, conformal prediction, etc. However, these methods are currently limited to the closed-end text generation domain, requires standardized text input/output formats and generation formats, and/or ground truth answers. Intuitively, these methods come from the statistical machine learning domain, and therefore are much better suited to e.g. tabular prediction tasks. Their application on LLMs is an adaptation rather than a new method overall. To the best of our knowledge, there is not a well-founded theory for SHAPLEY values in LLMs, especially the open-end text generation domain. On the other hand, there is one paper on the theory of conformal prediction in LLMs, but not related to perturbations \citep{mohri2024language}.

More specifically, feature attribution methods deserve its name because a score is assigned to each input feature, reflecting its impact on the generated model output. Early attribution methods that have been developed include (1) perturbation-based methods \citep{lundbergunified}, (2) gradient-based methods \citep{sundararajan2017axiomatic, montavon2017explaining}, (3) linear approximations \citep{ribeiro2016should}. Recently, these methods have been specifically adapted for LLMs \citep{sikdar2021integrated, enguehard2023sequential}. See \citep{singh2024rethinking} for a more complete literature review.

The gradient-inspired methods both break down the sentence into atomic elements (in language that would be tokens from byte pair encodings). Then, the gradient for each token can be calculated with respect to the final answer. This method suffers the same drawbacks as SHAPLEY values, specifically requiring standardized input/output text formats and/or ground truth answers. These papers demonstrate their method on closed-end tasks, such as text classification, prediction, etc. These methods were developed in tabular prediction tasks or compute vision tasks, and we have not done a really good job in adapting these methods to LLMs. An interesting parallel can be made that DBSA is attempting to adapt gradient-inspired methods on any open-end text generation.

\subsection{Fairness/Bias}
\label{sec:appendix-fairness/bias}

Certain methods in bias measurement and fairness quantification are also similar in spirit to DBSA. However, these works are limited to addressing specific subgroups, and while optional, they sometimes require human annotation (e.g. for bias annotations), which restricts their flexibility and applicability to arbitrary perturbations.

For example, DisCo \citep{webster2020measuring} is a seminal piece of work that proposes to look at model output distributions under perturbation in the prompt. To begin, each prompt template (e.g., “[X] is [MASK]”; “[X] likes to [MASK]”) has two slots. The first slot is manually filled with attributes associated with a social group (the original paper featured gendered names and nouns, but easily extended to other sensitive groups with well-defined word lists), and the second slot is filled by the model’s top three candidate predictions. The final score is the number of different predictions between social groups across all prompt templates. Bias is measured by the difference between normalized probability scores for two binary and opposing social group words. The main limitation of this approach is that it only addresses specific subgroups, and the output distribution is rather limited (closed-end generation). For a detailed discussion on this topic, see \citep{gallegos2024bias}.

Similar ideas have come up under works that investigate fairness under distributional shift. For example, \citep{wang2023robust} shows that while earlier methods proposed to adapt the model to be fair on the current known data distribution, or requires unlabeled data from the target distribution (assuming the target distribution is known), or requires the existence of a joint causal graph to represent the data distribution for all domains, their work aims to generalize fairness learned on current distribution to unknown and unseen target distributions. The field of fairness/bias also features the use of sensitivity analysis to explore how various aspects of the feature vector will affect a given outcome. Due to the introduction of SHAPLEY values, there has been an increase in attention from fairness researchers. \citep{cowgill2017algorithmic} proposed to perturb feature vectors to measure the effect on model performance of specific interventions. \citep{gao2015exploratory} investigated visual mechanisms to better display fairness/bias issues with data to users. \citep{jung2018algorithmic} used sensitivity analysis to evaluate sensitive variables and their relationships with classification outcomes, indicating that sensitivity analysis can help to better understand uncertainty with respect to fairness. However, these methods are subject to the same problems discussed before. Namely, they are not general enough in their input/output formats, and they sometimes require human annotation (e.g. for bias annotations).

On an \textit{application level}, we believe DBSA can directly improve the auditing of language modeling based systems to evaluate whether they are fair and reliable. That is, many applications today use language model as a system-level comonent for solving tasks. Consider a few examples. Language models are used to generate hypotheses to test existing ML systems \citep{rauba2024context}, improve information acquisition \citep{astorga2024active}, improve model robustness \citep{rauba2024self} or even improve model hyperparameters \citep{liu2024large}. Many such applications exist, too broad to be covered here \citep{hadi2023survey}. In most such works, the limitation section contains words of caution against fairness and bias---how can language models be trusted in such cases when they are used as external systems? Our work proposes a direct way to perform such audits.

\subsection{Direct Prompting}
\label{sec:appendix-direct-prompting}
Direct prompting methods generally work by either prompting along the lines of "Can you explain your logic" to an LLM, or in the case of Chain-of-Thought (CoT), goes back to changing specific values in the prompt in tasks like mathematical question-answering. In general, these methods either requires human intervention and examination, or ground truth answers labels, e.g. for process supervision, in context learning, RAG, etc. There has been efforts to make prompting methods gradient-inspired, see \citep{yuksekgonul2024textgrad} for an example.

Early adversarial attacks on LLMs apply simple character or token operations to trigger the LLM to generate incorrect predictions. Since these attacks usually generate misspelled prompts, they are easy to block in real-world applications. More recently, jailbreaking prompts are intentionally designed to bypass
the LLM built-in safeguard capabilities, eliciting the generation of harmful content. However, the discrete nature of text has significantly impeded learning more effective adversarial attacks against LLMs. Even for recent work that has developed gradient-based optimizers for efficient text modality attacks (\citep{wen2024hard} presented a gradient-based discrete optimizer that is suitable for attacking the text pipeline of CLIP), they still require access to a white-box language model, and is limited in their input/output formats.

\paragraph{Other interpretability methods.} We end the discussion on related works by documenting other interpretability methods that are also interesting approaches, but not immediately extended to the DBSA research. One promising method for understanding LLM representations is by looking at their attention heads \citep{clark2019does}, embeddings \citep{morris2023text}, and representations \citep{zou2023representation}. There are also methods that directly decode an output token to understand what is represented at different positions and layers \citep{belrose2023eliciting, ghandeharioun2024patchscope}. Finally, influence functions are interesting to study LLM behavior, but the idea of influence functions is limited to the impact on adding one example on the training set. Also, influence functions require the derivative, i.e. white-box access to LLMs. In general, methods in mechanistic interpretability (e.g. Attention head importance, circuit analysis, influence functions) are interesting approaches to address LLM interpretability, but none of them are easily extended to the DBSA framework, and we leave comparison to these methods for future work.

\newpage

\section{Implementation details}
\label{sec:implementation_details}

We provide a comprehensive guide for implementing Distribution-Based Sensitivity Analysis (DBSA). This section outlines the step-by-step process with specific code snippets and technical considerations.

\subsection{Prerequisites and Setup}
Our implementation uses Python 3.8 with NumPy, SciPy, Scikit-learn, and sentence-transformers libraries. We assume access to an LLM through a function \texttt{llm(prompt, n)} that generates \texttt{n} responses given a prompt.

\subsection{Text Preprocessing and Tokenization}
We tokenize the input text using a regular expression that captures words, numbers, and punctuation:

\begin{verbatim}
def tokenize_and_prepare_for_scoring(text):
    tokens = re.findall(r'\$?\d+(?:,\d+)*(?:\.\d+)?|\w+|[^\w\s]', text)
    token_positions = OrderedDict()
    for index, token in enumerate(tokens):
        if token not in token_positions:
            token_positions[token] = {'positions': [], 'score': None, 'pval': None}
        token_positions[token]['positions'].append(index)
    return tokens, token_positions
\end{verbatim}

\subsection{Sampling and Perturbing Responses}
\textbf{Key Point:} We sample multiple responses for both original and perturbed prompts to approximate their distributions.

\begin{verbatim}
def get_responses(prompt, n=40):
    return llm(prompt, n=n)

def perturb_sentence(tokens, perturb_index, neighbor):
    new_tokens = tokens.copy()
    new_tokens[perturb_index] = neighbor
    return new_tokens
\end{verbatim}

There are many ways to generate perturbations. In our example, we generate perturbations by asking an external LLM to generate $k$ synonyms. Other possibilities include relying on external libraries, e.g. word2vec.

\textbf{Embeddings}. We use the OpenAI sentence transformer model ''text-embedding-ada-002'' to generate embeddings.

\subsection{Computing Energy Distance}
\textbf{Key Point:} We use energy distance as our metric for comparing distributions.

\begin{verbatim}
def compute_energy_distance(X, Y, distance='cosine'):
    n, m = len(X), len(Y)
    dists_XY = cdist(X, Y, distance)
    dists_XX = cdist(X, X, distance)
    dists_YY = cdist(Y, Y, distance)
    
    term1 = (2.0 / (n * m)) * np.sum(dists_XY)
    term2 = (1.0 / n**2) * np.sum(dists_XX)
    term3 = (1.0 / m**2) * np.sum(dists_YY)
    
    return term1 - term2 - term3
\end{verbatim}

\subsection{Performing Statistical Tests}
We use a permutation test to calculate p-values:

\begin{verbatim}
def permutation_test_energy(X, Y, num_permutations=500, distance='cosine'):
    combined = np.vstack((X, Y))
    n = len(X)
    E_values = []
    for _ in range(num_permutations):
        np.random.shuffle(combined)
        perm_X, perm_Y = combined[:n], combined[n:]
        E_perm = compute_energy_distance(perm_X, perm_Y, distance)
        E_values.append(E_perm)
    return np.array(E_values)

def compute_energy_distance_fn(baseline_embeddings, perturbed_embeddings):
    E_n = compute_energy_distance(baseline_embeddings, perturbed_embeddings)
    E_values = permutation_test_energy(baseline_embeddings, perturbed_embeddings)
    p_value = np.mean(E_values >= E_n)
    return E_n, p_value
\end{verbatim}

\subsection{Alternative Implementations}
Alternatively, we could combine the distance calculation + statistical test into one step:

\begin{verbatim}
def calculate_difference_and_pvalue(arr1, arr2, num_permutations=1000, mode="energy"):
    if mode == "mean":
        observed_diff = np.abs(np.mean(arr1) - np.mean(arr2))
    elif mode == "EMD":
        observed_diff = wasserstein_distance(arr1, arr2)
    elif mode == "energy":
        observed_diff = energy_distance(arr1, arr2)
    else:
        raise ValueError(f"Invalid mode: {mode}")
    combined = np.concatenate([arr1, arr2])
    n1, n2 = len(arr1), len(arr2)
    permutation_diffs = []
    for _ in range(num_permutations):
        np.random.shuffle(combined)
        permuted_arr1, permuted_arr2 = combined[:n1], combined[n2:]
        if mode == "mean":
            permuted_diff = np.mean(permuted_arr1) - np.mean(permuted_arr2)
        elif mode == "EMD":
            permuted_diff = wasserstein_distance(permuted_arr1, permuted_arr2)
        elif mode == "energy":
            permuted_diff = energy_distance(permuted_arr1, permuted_arr2)
        else:
            raise ValueError(f"Invalid mode: {mode}")
        permutation_diffs.append(permuted_diff)
    p_value = np.mean(np.abs(permutation_diffs) >= np.abs(observed_diff))
    return observed_diff, p_value
\end{verbatim}

\textbf{Optimization}. For improved efficiency, especially with longer texts, we implement parallel processing using Python's \texttt{multiprocessing} module to distribute token perturbation and analysis across available CPU cores.

\newpage

\newpage
\section{Extended experiments}
\label{sec:extended_experiments}
\textbf{Purpose}. This section provides additional insights into the similarity between choosing different distance functions for the energy-based calculations across different models. We run each model with different distances and evaluate the Spearman rank correlation between the given token outputs, where 1 indicates perfect ranking-based correlaction and 0 indicates no relationship.

\textbf{Discussion}. Across all models, we find that the distance metrics consistently rank the same words as important for DBSA. 

\begin{figure}[htbp]
    \centering
    \begin{multicols}{3}
        \begin{subfigure}[b]{0.3\textwidth}
            \centering
            \includegraphics[width=\linewidth]{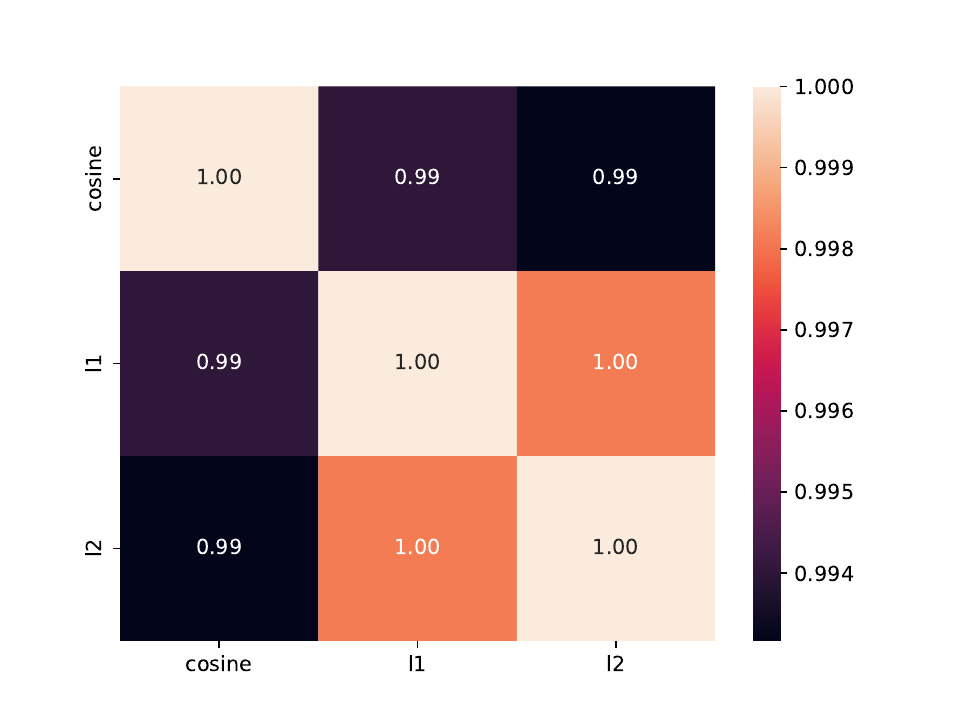}
            \caption{GPT-3.5}
        \end{subfigure}
        
        \begin{subfigure}[b]{0.3\textwidth}
            \centering
            \includegraphics[width=\linewidth]{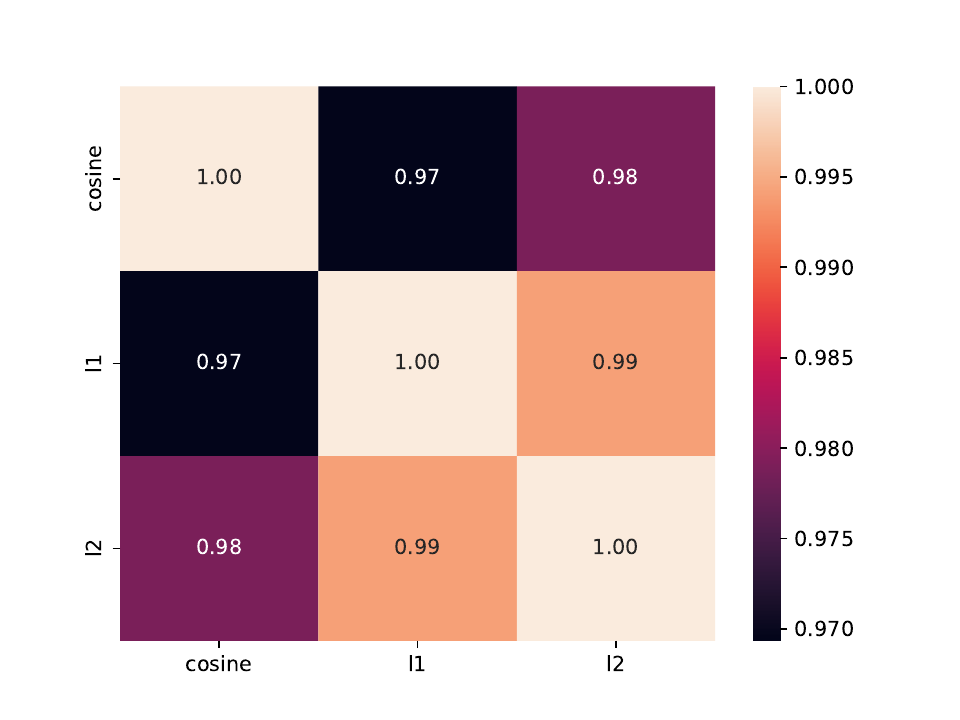}
            \caption{GPT-4}
        \end{subfigure}
        
        \begin{subfigure}[b]{0.3\textwidth}
            \centering
            \includegraphics[width=\linewidth]{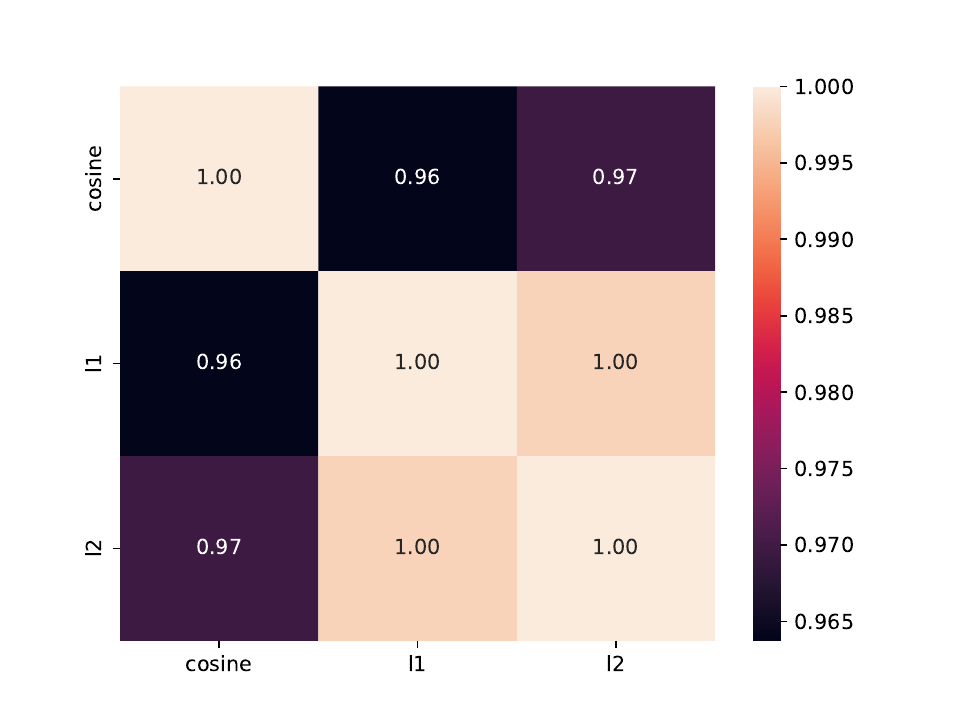}
            \caption{SmolLM-135M}
        \end{subfigure}
        
        \begin{subfigure}[b]{0.3\textwidth}
            \centering
            \includegraphics[width=\linewidth]{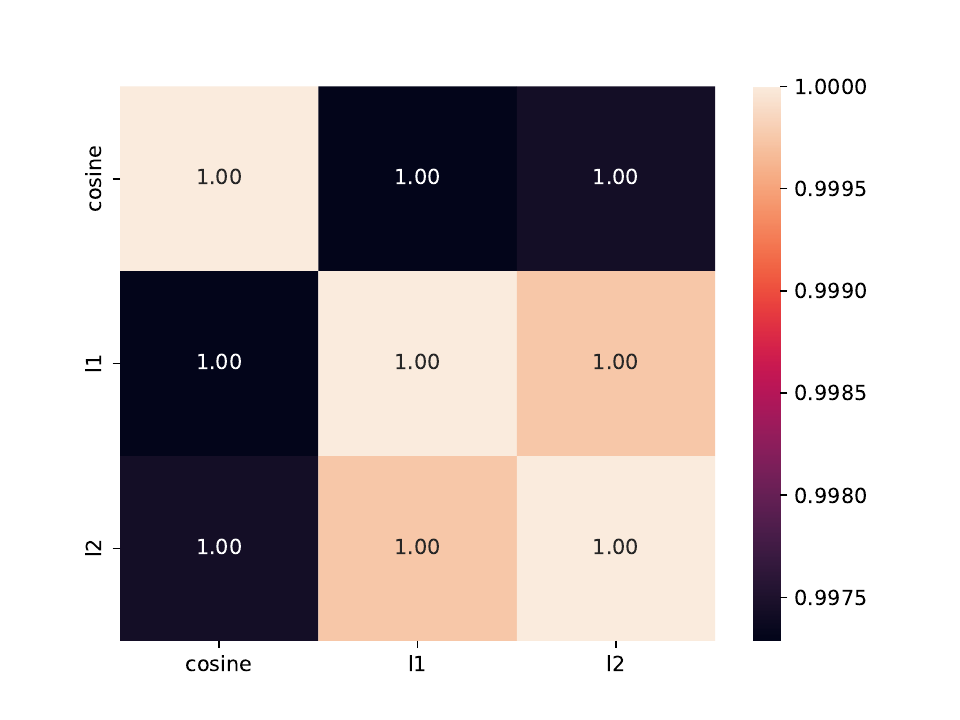}
            \caption{MagicPrompt-Stable-Diffusion}
        \end{subfigure}
        
        \begin{subfigure}[b]{0.3\textwidth}
            \centering
            \includegraphics[width=\linewidth]{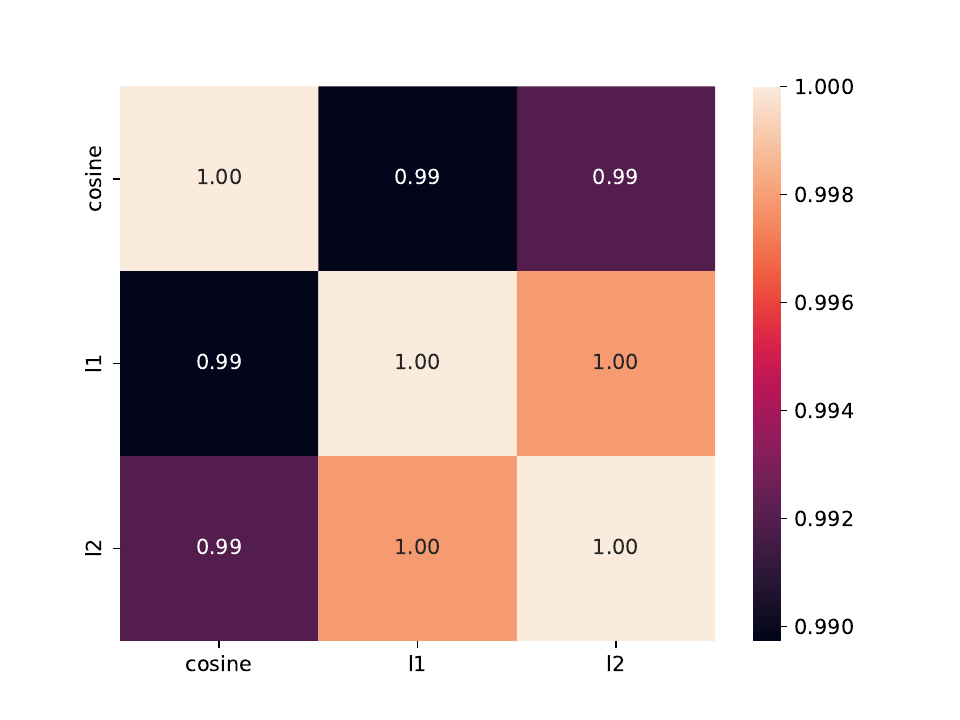}
            \caption{Phi-3-mini-4k-instruct}
        \end{subfigure}
        
        \begin{subfigure}[b]{0.3\textwidth}
            \centering
            \includegraphics[width=\linewidth]{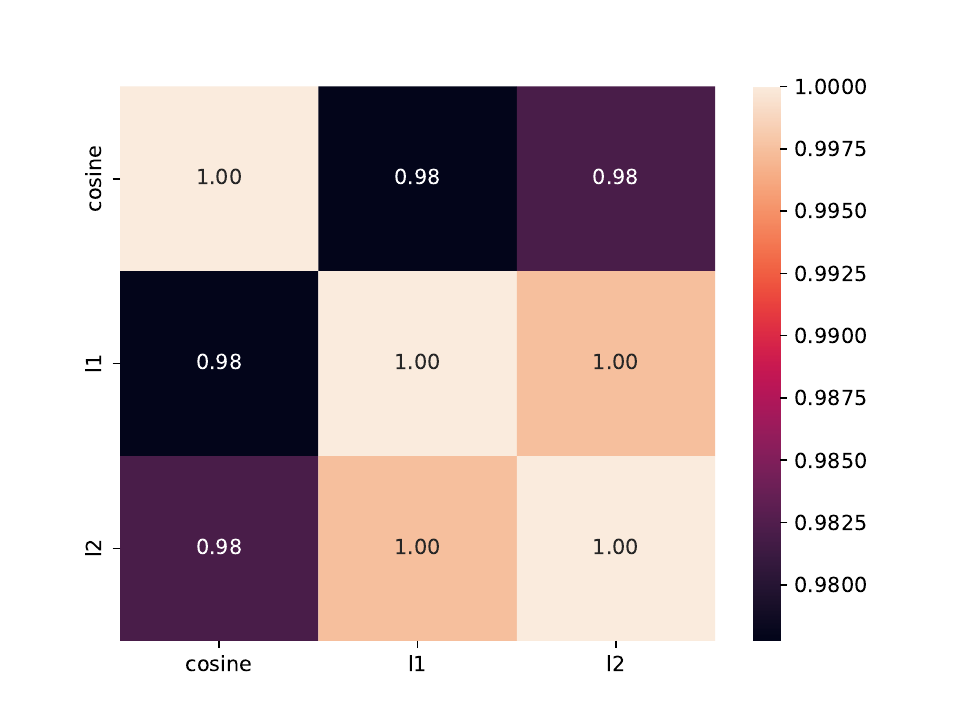}
            \caption{Mistral-7B-Instruct-v0.2}
        \end{subfigure}
        
        \begin{subfigure}[b]{0.3\textwidth}
            \centering
            \includegraphics[width=\linewidth]{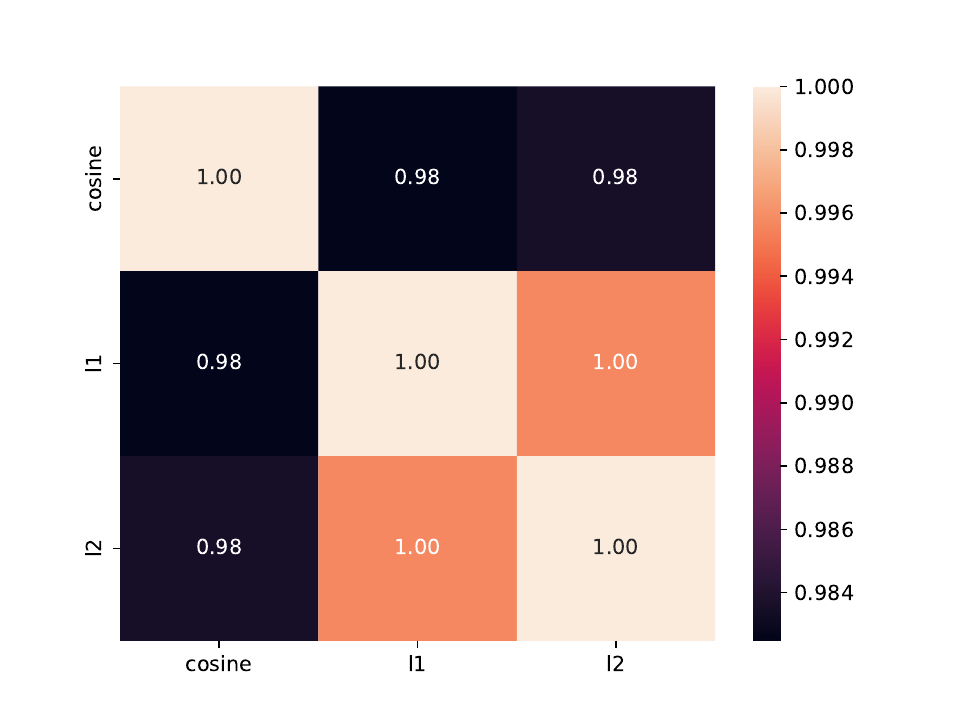}
            \caption{Meta-Llama-3.1-8B-Instruct}
        \end{subfigure}
        
        \begin{subfigure}[b]{0.3\textwidth}
            \centering
            \includegraphics[width=\linewidth]{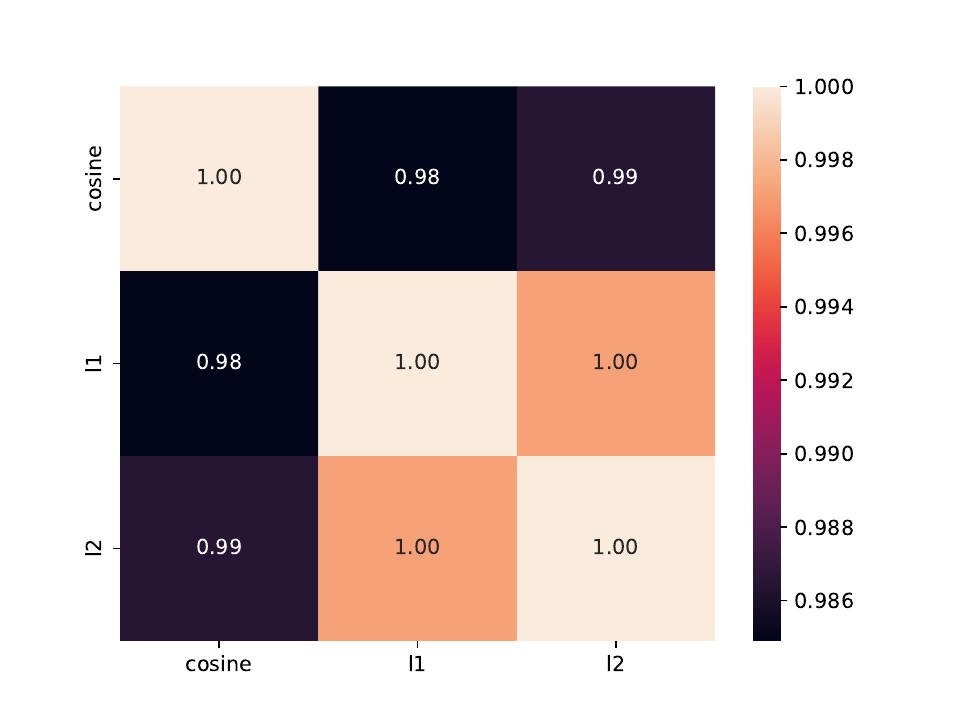}
            \caption{gemma-2-9b-it}
        \end{subfigure}
    \end{multicols}
    \caption{Similarity functions for various models}
    \label{fig:similarity_functions}
\end{figure}

\newpage

\section{Experimental Details}
\label{sec:experimental_setting}

\subsection{Algorithm}

The following is an implementation of DBSA that is used for evaluating the sensitivity of each word. We use the same notation as in the main part of the paper.

\begin{algorithm}[H]
\caption{Distribution-Based Sensitivity Analysis}
\label{alg:sensitivity}
\begin{algorithmic}[1]
\REQUIRE Input sequence $x = (w_1, \ldots, w_n)$; sample sizes $n, m$; nearest neighbors $k$; embedding function $\phi: \mathcal{Y} \rightarrow \mathbb{R}^d$; similarity function $s: \mathbb{R}^d \times \mathbb{R}^d \rightarrow \mathbb{R}$; distance function $D: \mathbb{R}^d \times \mathbb{R}^d \rightarrow \mathbb{R}$
\ENSURE Sensitivity scores $\{ E_w, p_w \}_{w \in T}$
\STATE $T \gets \{ w \mid w \in x \}$ $\quad$ \text{ // \textit{Unique tokens}}
\FORALL{$w \in T$}
    \STATE $\mathcal{I}_w \gets \{ i \mid t_i = w \}$ $\quad$ \text{ //\textit{ Token positions}}
    \FORALL{$i \in \mathcal{I}_w$}
        \STATE $\mathcal{N}_k(t_i) \gets \argmin_{w' \in \mathcal{X}, w' \neq t_i} \{ D(\phi(w'), \phi(t_i))\}_{k}$ 
        \FORALL{$t_i' \in \mathcal{N}_k(t_i)$}
            \STATE $x^{(t_i')} \gets (w_1, \ldots, w_{i-1}, t_i', w_{i+1}, \ldots, w_n)$
            \STATE $Y \gets \{ y^{(j)} \}_{j=1}^n \sim \mathcal{S}(x)$, $Y' \gets \{ y'^{(j)} \}_{j=1}^m \sim \mathcal{S}(x^{(t_i')})$
            \STATE $\Phi \gets \{ \phi(y^{(j)}) \}_{j=1}^n$, $\Phi' \gets \{ \phi(y'^{(j)}) \}_{j=1}^m$
            \STATE $A_{wi}^{(w'_i)} \gets \frac{2}{nm} \sum_{a=1}^n\sum_{b=1}^m s(\phi_a, \phi'_b)$
            \STATE $B_{wi}^{(w'_i)} \gets \frac{1}{n^2} \sum_{a=1}^n\sum_{b=1}^n s(\phi_a, \phi_b)$
            \STATE $C_{wi}^{(w'_i)} \gets \frac{1}{m^2} \sum_{a=1}^m\sum_{b=1}^m s(\phi'_a, \phi'_b)$
            \STATE $E_{wi}^{(w'_i)} \gets A_{wi}^{(w'_i)} - B_{wi}^{(w'_i)} - C_{wi}^{(w'_i)}$
            \STATE $p_{wi}^{(w'_i)} \gets \text{PermutationTest}(\Phi \cup \Phi', E_{wi}^{(w'_i)})$ 
        \ENDFOR
        \STATE $\bar{E}_{wi} \gets \frac{1}{k} \sum_{t_i' \in \mathcal{N}_k(t_i)} E_{wi}^{(w'_i)}, \bar{p}_{wi} \gets \frac{1}{k} \sum_{t_i' \in \mathcal{N}_k(t_i)} p_{wi}^{(w'_i)}$
    \ENDFOR
    \STATE $E_w \gets \frac{1}{|\mathcal{I}_w|} \sum_{i \in \mathcal{I}_w} \bar{E}_{wi}, p_w \gets \frac{1}{|\mathcal{I}_w|} \sum_{i \in \mathcal{I}_w} \bar{p}_{wi}$
\ENDFOR
\RETURN $\{ E_w, p_w \}_{w \in T}$
\end{algorithmic}
\end{algorithm}

\subsection{Language models used}

\paragraph{Compute Resources.} All the experiments in this paper were carried out on an A100 machine, with NVIDIA driver version 535.183.06 and CUDA12. The GPU VRAM is 80GB.

\paragraph{Language Models.} A total of 8 language models were used for the experiments of this paper

\begin{itemize}
    \item \textbf{GPT-3.5.} The GPT-3.5 model is a deployment taken from OpenAI endpoints. The deployment version is gpt-35-1106.
    \item \textbf{GPT-4.} The GPT-4 model is a deployment taken from OpenAI endpoints. The deployment version is gpt-4-0613-20231016.
    \item \textbf{SmolLM-135M.} The SmolLM-135M model is taken from Huggingface. The full model specification is HuggingFaceTB/SmolLM-135M.
    \item \textbf{MagicPrompt-Stable-Diffusion.} The MagicPrompt-Stable-Diffusion model is taken from Huggingface. The full model specification is Gustavosta/MagicPrompt-Stable-Diffusion.
    \item \textbf{Phi-3-mini-4k-instruct.} The Phi-3-mini-4k-instruct model is taken from Huggingface. The full model specification is microsoft/Phi-3-mini-4k-instruct.
    \item \textbf{Mistral-7B-Instruct-v0.2.} The Mistral-7B-Instruct-v0.2 model is taken from Huggingface. The full model specification is mistralai/Mistral-7B-Instruct-v0.2.
    \item \textbf{Meta-Llama-3.1-8B-Instruct.} The Meta-Llama-3.1-8B-Instruct model is taken from Huggingface. The full model specification is meta-llama/Meta-Llama-3.1-8B-Instruct.
    \item \textbf{gemma-2-9b-it.} The gemma-2-9b-it model is taken from Huggingface. The full model specification is google/gemma-2-9b-it.
\end{itemize}

\subsection{Hyperparameter details}

In our experiments with DBSA, we conducted ablations for most hyperparameters, as detailed in the main paper. However, it's challenging to be fully exhaustive. The following list outlines the key hyperparameters and considerations for each step of the DBSA process:

\begin{enumerate}
    \item \textbf{LLM Sampling:} This step involves standard LLM sampling hyperparameters. We set the temperature to 1 and the max-length to 256. For more details on these and other sampling parameters, refer to the Huggingface documentation.
    
    \item \textbf{Perturbation Generation:} This step requires selecting a perturbation method (e.g., synonyms, antonyms) and a function to generate these perturbations. In our experiments, we defined sensitivity as nearest-neighbor sensitivity to approximate gradients of the input prompt. We used GPT-4 to generate synonyms for each word, finding it empirically superior to alternatives like word2vec.
    
    \item \textbf{Perturbed Prompt Sampling:} This step uses the same hyperparameters as Step 1.
        
    \item \textbf{Embedding:} We used OpenAI's Ada embeddings, as they effectively capture meaningful semantic information in the prompts.
    
    \item \textbf{Distance Calculation:} The selection of a distance function is key to measuring the disparity between response embeddings. Our experiments showed that cosine, L1, and L2 distances yielded similar results. We opted for cosine distance in our final experiments.
    
    \item \textbf{Statistical Analysis:} The choice of statistical metric for calculating effect size between distributions during permutation testing is important. We selected the energy distance as our metric for effect size.
\end{enumerate}

\newpage

\section{Prompt examples}
\label{sec:prompt_examples}

\subsection{Prompts}
The following prompts were the primary prompts used to evaluate DBSA sensitivity.

\noindent
\begin{minipage}{\linewidth}
\captionof{lstlisting}{Prompt example 1}
\begin{lstlisting}[language=Python]
text_legal = """Company A agrees to pay Company B $10 million for developing a revolutionary AI software within 12 months. If Company B fails to deliver a fully functional product by the deadline, they must refund 50% of the payment and provide an additional 3 months of development at no extra cost. However, if the delay is due to circumstances beyond Company B's reasonable control, these penalties shall not apply. This agreement is governed by California law and any disputes shall be resolved through binding arbitration."""
\end{lstlisting}
\end{minipage}

\vspace{1em}

\noindent
\begin{minipage}{\linewidth}
\captionof{lstlisting}{Prompt example 2}
\begin{lstlisting}[language=Python]
text_medical = """Patient is a 45-year-old male presenting with progressive dyspnea on exertion over the past two weeks. On examination, patient appears mildly distressed. Lower extremities show 2+ pitting edema to mid-shin bilaterally. Chest X-ray shows pulmonary vascular congestion. Clinical presentation is consistent with new-onset congestive heart failure, likely due to hypertensive heart disease."""
\end{lstlisting}
\end{minipage}

\noindent
\begin{minipage}{\linewidth}
\captionof{lstlisting}{Prompt example 3}
\begin{lstlisting}[language=Python]
text_trading = """A senior executive is accused of insider trading , allegedly using confidential information to gain substantial financial benefits . The defendant maintains his innocence , claiming that all investment decisions were based on public data ."""
\end{lstlisting}
\end{minipage}

\noindent
\begin{minipage}{\linewidth}
\captionof{lstlisting}{Prompt example 4}
\begin{lstlisting}[language=Python]
text_manufacturing = A manufacturing company was sued for producing a faulty product that caused significant injuries to a customer ."""
\end{lstlisting}
\end{minipage}

\newpage
\subsection{Example nearest neighbors}
This section outlines example nearest neighbors for different tokens.

\noindent
\begin{minipage}{\linewidth}
\captionof{lstlisting}{Closest words for prompt 3}
\begin{lstlisting}[language=Python]
closest_words = {"Defendant": ["Accused", "Respondent", "Litigant"], ".": [",", "!", "?"], ",": [";", ".", ":"], "Senior": ["Top-tier", "High-ranking", "Upper-level"], "Executive": ["Administrator", "Officer", "Manager"], "Accused": ["Alleged", "Charged", "Indicted"], "Insider": ["Internal", "In-house", "Privileged"], "Trading": ["Dealing", "Stock-jobbing", "Market manipulation"], "Allegedly": ["Supposedly", "Reportedly", "Purportedly"], "Using": ["Utilizing", "Employing", "Applying"], "Confidential": ["Private", "Secret", "Classified"], "Information": ["Data", "Details", "Intelligence"], "To": ["Toward", "For", "In order to", "So as to"], "Gain": ["Acquire", "Secure", "Obtain"], "Substantial": ["Significant", "Considerable", "Major"], "Financial": ["Monetary", "Fiscal", "Economic"], "Benefits": ["Advantages", "Gains", "Profits"], "Maintains": ["Affirms", "Asserts", "Insists"], "Innocence": ["Guiltlessness", "Blamelessness", "Purity"], "Claiming": ["Asserting", "Stating", "Contending"], "Investment": ["Financial", "Capital", "Asset"], "Decisions": ["Choices", "Determinations", "Conclusions"], "Public": ["Open", "Publicly available", "Common"], "Data": ["Information", "Statistics", "Facts"], "A": ["An", "One", "Any"], "Is": ["Exists", "Stands", "Remains", "Constitutes"], "Of": ["Concerning", "Regarding", "About", "Pertaining to"], "Using": ["Utilizing", "With the help of", "Via", "By means of"], "To": ["Toward", "For", "In order to", "So as to"], "The": ["This", "That", "Said"], "His": ["Their", "Its", "Her"], "That": ["Which", "This", "What"], "All": ["Every", "Each", "Any"], "Were": ["Had been", "Were being", "Used to be"], "Based": ["Founded", "Established", "Built", "Grounded"], "On": ["Upon", "Over", "About", "Concerning"]}

\end{lstlisting}
\end{minipage}

\noindent
\begin{minipage}{\linewidth}
\captionof{lstlisting}{Closest words for prompt 4}
\begin{lstlisting}[language=Python]
closest_words = {"Manufacturing": ["Production", "Industrial", "Fabrication"], "Company": ["Corporation", "Firm", "Business"], "Was": ["Had been", "Was being", "Were"], "Sued": ["Prosecuted", "Litigated against", "Indicted"], "For": ["Because of", "Due to", "On account of"], "Producing": ["Creating", "Making", "Generating"], "A": ["One", "An", "Any"], "Faulty": ["Defective", "Damaged", "Malfunctioning"], "Product": ["Good", "Item", "Commodity"], "That": ["Which", "Who", "That which"], "Caused": ["Provoked", "Led to", "Resulted in"], "Significant": ["Major", "Considerable", "Substantial"], "Injuries": ["Harm", "Damage", "Trauma"], "To": ["Towards", "In relation to", "With respect to"], "A": ["One", "An", "Any"], "Customer": ["Consumer", "Client", "Purchaser"], ".": [",", "!", "?"]}
\end{lstlisting}
\end{minipage}

\end{document}